\documentclass[journal]{IEEEtran}

\usepackage{graphicx} % Required for inserting images
\usepackage{import} % for importing pdf_tex images
\usepackage{cite}
\usepackage{balance}          % to balance columns at the end
\usepackage{amsmath,amsfonts}
\usepackage{amsthm} % for proofs to work
\usepackage{algorithm}
\usepackage{algpseudocode}
\usepackage{units} % \unit
\usepackage{subcaption} % \subfigure
\usepackage{xfrac} % \sfrac
\usepackage{hhline}
\usepackage{booktabs}
\usepackage{mathtools} % \coloneq
\usepackage{array} % \tabular m{} option
\usepackage{ifthen} % \ifthenelse
\usepackage{hyperref} % \url
\usepackage{cleveref} % \cref
\usepackage{color} % for including colored fonts

\renewcommand{\paragraph}[1]{\noindent\textbf{#1}.}

\newcounter{factorcounter}[subsubsection]

\newcommand{\storeheadertext}[2]{%
    \expandafter\gdef\csname storedheadertext@#1\endcsname{#2}%
}

\newcommand{\getstoredheadertext}[1]{%
  \csname storedheadertext@#1\endcsname%
}

\newenvironment{factor}[2]{%
    \refstepcounter{factorcounter}% Increment the counter
    \storeheadertext{\thefactorcounter}{#1} % Store header text for the current counter
    \subsubsection*{#1~Factor~--~#2}
}{%
}

\Crefformat{factorcounter}{#2\getstoredheadertext{#1}~Factor#3}
\Crefmultiformat{factorcounter}{#2\getstoredheadertext{#1}#3}{ and~#2\getstoredheadertext{#1} Factors#3}{, #2\getstoredheadertext{#1}#3}{ and~#2\getstoredheadertext{#1} Factors#3}

\Crefname{appdx}{Appendix}{Appendices}

\newcommand{\ASDFp}{ASDF\textsuperscript{+}}
\newcommand{\ASDFm}{ASDF\textsuperscript{--}}

\algnewcommand\Not{\textbf{not}}

\newtheorem{theorem}{Theorem}

\title{Adaptive Self-Calibration for Minimalistic Collective Perception by Imperfect Robot Swarms}

\author{Khai~Yi~Chin,~\IEEEmembership{Graduate~Student~Member,~IEEE,} and Carlo~Pinciroli,~\IEEEmembership{Member,~IEEE}%
\thanks{This paper has supplementary downloadable material available at \url{http://ieeexplore.ieee.org}, provided by the authors. This includes all the HDF data files containing the convergence and accuracy data discussed in this manuscript and Jupyter notebooks to parse the data files. This material is 81.3 MB in size.}
\thanks{K.~Y.~Chin and C.~Pinciroli are with the NEST Lab at Worcester Polytechnic Institute, Worcester, MA 01609, USA (email: kchin@wpi.edu, cpinciroli@wpi.edu).}}

\begin{document}

\maketitle

\begin{abstract}
    Collective perception is a fundamental problem in swarm robotics, often cast as best-of-$n$ decision-making. Past studies involve robots with perfect sensing or with small numbers of faulty robots. We previously addressed these limitations by proposing an algorithm, here referred to as Minimalistic Collective Perception (MCP)~\cite{chinMinimalisticCollectivePerception2023}, to reach correct decisions despite the entire swarm having severely damaged sensors. However, this algorithm assumes that sensor accuracy is known, which may be infeasible in reality. In this paper, we eliminate this assumption to \textit{(i)} investigate the decline of estimation performance and \textit{(ii)} introduce an Adaptive Sensor Degradation Filter (ASDF) to mitigate the decline. We combine the MCP algorithm and a hypothesis test to enable adaptive self-calibration of robots' assumed sensor accuracy. We validate our approach across several parameters of interest. Our findings show that estimation performance by a swarm with correctly known accuracy is superior to that by a swarm unaware of its accuracy. However, the ASDF drastically mitigates the damage, even reaching the performance levels of robots aware \textit{a priori} of their correct accuracy.
\end{abstract}
\section{Introduction}
\label{sec:introduction}

\IEEEPARstart{C}{ollective} perception is a fundamental problem in autonomous robotics that forms the base of numerous robotic applications. Beyond localization, a robot's awareness of its surroundings is useful in generating a representation of the environment. Some example applications include the assessment of structural surface conditions~\cite{siemensmaCollectiveBayesianDecisionMaking2024,haghighatApproachBasedParticle2022} or the estimation of pollution sources~\cite{tamjidiUnifyingConsensusCovariance2021}. In these situations, leveraging a swarm robotic system can overcome prohibitive costs in time and resources that accompany single-robot solutions: a lone robot is likely slower to complete such areal coverage tasks and could demand more resources to operate. In particular, the numeric advantage of a robot swarm promises the benefits of robustness and scalability. Each member uses local information---much like in biological swarms---to collectively accomplish a task. In our case, the task of interest is to create a coherent representation of the environment. Then, not only should each robot's estimation of the environment be accurate, but the swarm needs to reach a consensus on said estimates; this is the \emph{collective perception} problem.

In this paper, we study collective perception using swarm robots with \emph{imperfect} sensors. The sensors are imperfect because they have a significant probability to report incorrect observations, for reasons such as sensor noise, faults, or due to an ill-trained neural network classifier~\cite{majcherczykDistributedDataStorage2021}. Aside from our previous work~\cite{chinMinimalisticCollectivePerception2023}, past research on collective perception broadly assumes robots to have perfect or near-perfect sensing, making the study of an imperfect swarm a novelty despite its apparent practicality. We consider an environment with a binary feature based on the scenario of Valentini \textit{et al.}~\cite{valentiniCollectivePerceptionEnvironmental2016}. Our robots operate in an arena of black-and-white tiles to collectively estimate the average number of black tiles (defined as the \emph{fill ratio}), \textit{i.e.,} the frequency at which a binary environmental feature occurs. The estimated parameter is used to decide whether the environment has more black or white tiles. We have previously shown success in doing so with an imperfect swarm using an estimation algorithm~\cite{chinMinimalisticCollectivePerception2023}. In this manuscript, we shall refer to our aforementioned algorithm as the Minimalistic Collective Perception (MCP) algorithm. However, this algorithm requires the exact sensor accuracy levels to be known by the robots. While not an unreasonable assumption (attainable through calibration, for example), this is limiting in instances where robot repairs are costly or infeasible. For that reason, we investigate the effect of breaking this assumption and propose an approach to rectify incorrectly assumed sensor accuracies. That is, a robot's assumed sensor accuracy---which may differ from the true accuracy---now becomes part of the original estimation problem (on top of the environmental parameter). Note that we are primarily concerned with the swarm's fill ratio estimate in this work---as opposed to their converted binary decision as is often done~\cite{shanDiscreteCollectiveEstimation2021,ebertBayesBotsCollective2020,crosscombeRobustDistributedDecisionmaking2017}---since that ultimately dictates the decision-making quality.

\begin{figure*}[t]
    \centering
    \def\svgwidth{\textwidth}
    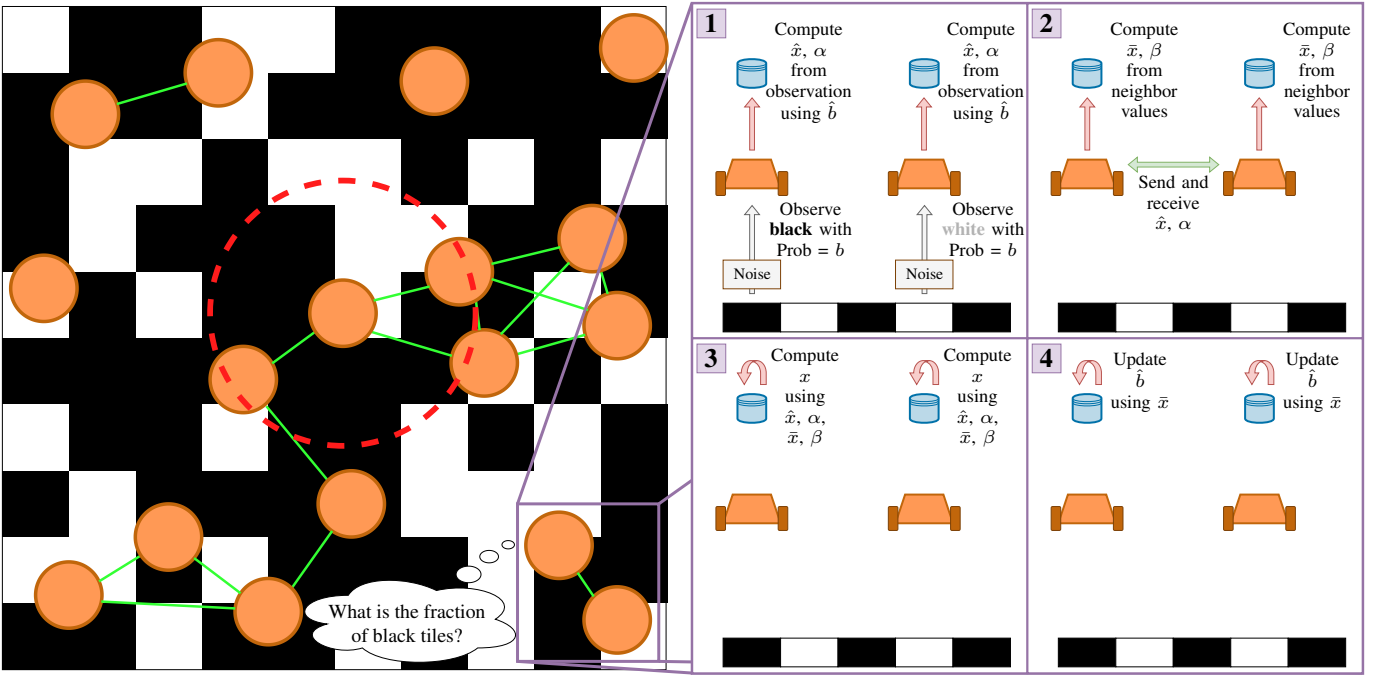
    \caption{A top-down view of a robotic swarm (orange circles) in a black-and-white tile environment with a focus on the interaction between two robots. Communication links between robots (green lines) are established if neighbors enter the communication range of a robot (red dotted line, drawn for a single robot). The expanded panels show the two robots using the MCP algorithm to perform collective perception (steps 1-3), in which the assumed accuracy information is obtained with the ASDF (step 4).}
    \label{fig:banner_image}
\end{figure*}

Our main contribution is the introduction of an Adaptive Sensor Degradation Filter (ASDF) that enables the swarm robots to self-calibrate. The ASDF adjusts a robot's assumed accuracy upon detecting a difference---using a hypothesis test based on a $t$-test statistic---between said robot's local estimate and its neighbors'. We extend the MCP algorithm with the ASDF by providing an updated assumed accuracy as the last component, as shown in \Cref{fig:banner_image}: \textit{(i)} Each robot computes a \emph{local estimate} (that relies on an assumed accuracy) based on a local observation of the environment; \textit{(ii)} The robots exchange local estimates between themselves and compute a \emph{social estimate} from the received estimates; \textit{(iii)} An \emph{informed estimate} is computed by each robot as a weighted average of its local and social estimates; \textit{(iv)} Each robot \emph{adaptively adjusts its assumed accuracy} based on the social estimate. This combined algorithm thus provides an informed estimate that serves as a robot's best guess of the environmental feature, without the need of a correct assumed accuracy. Moreover, the minimal requirements of the MCP are preserved with our proposed approach: $O(1)$ for computational capabilities and bandwidth and $O(q)$ for storage of the $t$-test table with $q$ elements.

To evaluate the effectiveness of the ASDF, we ran experiments across different proportions of robots with incorrect assumed accuracies and filter activation periods under varying swarm densities, assumed-true accuracy deviations, and fill ratios. We find that the ASDF drastically mitigates the decline in estimation performance by a swarm unaware of its sensor accuracy (\textit{i.e.}, a flawed swarm) when compared to a swarm with correctly known accuracy (\textit{i.e.}, a flawless swarm) even as the flawed robots are no longer a minority. That said, in the extreme case where $100\%$ of the swarm has a flawed assumed accuracy, we do observe limitations to the effectiveness of the ASDF. The remainder of the paper is organized as follows. \Cref{sec:related_work} surveys the related work and \Cref{sec:methodology} describes the derivation of the MCP and ASDF algorithms, while \Cref{sec:simulated_experiments} and \Cref{sec:physical_experiments} discuss the simulated and physical experimental results respectively. We conclude the paper in \Cref{sec:conclusions}.
\section{Related Work}
\label{sec:related_work}

Research on collective perception in robot swarms is extensive. The most influential approaches are based on data-driven techniques or are derived from mathematical first principles. Often, collective perception is studied as a best-of-$n$ decision-making problem, where robots draw on the perceived environmental attributes to make decisions. Some instances include judging the health of a structural entity \cite{siemensmaCollectiveBayesianDecisionMaking2024} or deciding on the correct set of annotations on surrounding objects \cite{majcherczykDistributedDataStorage2021}. In this work, the decision concerns the selection of the most frequent of two colors, which is an abstraction for site-selection problems. Reina \textit{et al.}~\cite{reinaDesignPatternDecentralised2015} formalized an approach that accounts for micro- and macroscopic behaviors for a site-selection task. Likewise, Schmickl \textit{et al.}~\cite{schmicklCollectivePerceptionRobot2007} proposed a honeybee-inspired approach for robots to select the best of two sites. Valentini \textit{et al.}~\cite{valentiniCollectivePerceptionEnvironmental2016} also studied a best-of-$2$ decision-making problem where robots decide the most frequent tile color in a black-and-white tile environment. The authors introduced the majority and voter models that are often used by later works to benchmark collective perception algorithms. One such work is by Almansoori \textit{et al.}~\cite{almansooriComparativeStudyDecision2022}, who compared the voter model against their proposed recurrent neural-network-based algorithm. Probabilistic approaches have also been explored using frequentist~\cite{abdelliMaximumLikelihoodEstimate2023} and Bayesian methods~\cite{ebertBayesBotsCollective2020,shanDiscreteCollectiveEstimation2021}. Our study, like the ones discussed so far, focuses on a static environment, \textit{i.e.,} the frequency of the tile colors remains the same. Some studies have instead explored the collective perception problem using a dynamic environment like those by Pfister \textit{et al.}~\cite{pfisterCollectiveDecisionmakingBayesian2022,pfisterCollectiveDecisionmakingChange2023}.

The works above consider swarm robots to have perfect sensing, which may be unrealistic. To that end, Morlino \textit{et al.}~\cite{morlinoCollectivePerceptionSwarm2010} synthesized controllers for collective perception with noisy sensors, though robots are able to communicate globally. Crosscombe \textit{et al.}~\cite{crosscombeRobustDistributedDecisionmaking2017} investigated a 3-state voter model within a swarm that contains a subset of unreliable robots. When robots are outright malicious, Strobel \textit{et al.}~\cite{strobelManagingByzantineRobots2018,strobelBlockchainTechnologySecures2020,strobelRobotSwarmsNeutralize2023} proposed blockchain-based methods to enable swarm consensus in the presence of Byzantine robots. Such methods are computationally demanding, however. With minimalism in mind, we presented the MCP algorithm that is robust to an entire swarm of imperfect robots~\cite{chinMinimalisticCollectivePerception2023}. MCP is effective even for robots with severe sensor degradation (where the readings are wrong up to $45\%$ of the time), though a crucial assumption is that the robots' true sensor accuracy is known.

In this work, we analyze the performance of MCP with no knowledge of the true sensor accuracy and introduce the ASDF to mitigate the resulting adverse effects. Our contribution is akin to an endogenous, model-based fault mitigation mechanism, aimed at the robots' assumed sensor accuracies. Recent work on swarm faults include detection~\cite{christensenExogenousFaultDetection2007,khadidosExogenousFaultDetection2015,taraporeFaultDetectionSwarm2019} and diagnosis~\cite{okeeffeFaultDiagnosisRobot2017,okeeffeFaultDiagnosisRobot2017a} methods, but mitigation methods---like ours---remain model- and application-specific. For example, Azizi \textit{et al.}~\cite{aziziHierarchicalArchitectureCooperative2012} designed a framework for distributed satellite formations that allow recovery procedures, although all members need to be at least connected to one other member. Similarly, Taheri \textit{et al.}~\cite{taheriCyberattackMachineinducedFault2024} studied a distributed fault detection methodology for cyberphysical systems, albeit within the context of isolating cyberattacks.

Outside distributed systems, sensor degradation estimation is an active field of research in manufacturing and industrial engineering. Elwany and Gebraeel~\cite{elwanyRealTimeEstimationMean2009} proposed a Bayesian approach to estimate the remaining life distributions of degraded sensors based on an inverse Gaussian distribution. Other Bayesian-based estimators were studied by Liu~\textit{et al.}~\cite{liuStochasticFilteringApproach2020} (Kalman filter) to design a condition-based maintenance policy for sensors, and by Liu~\textit{et al.}~\cite{liuNovelAlgorithmQuantized2023} (particle filter) to track a target using multiple degraded sensors. Unlike those works, we consider sensor degradation estimation within the context of a decentralized swarm of robots. The robots in our approach estimate the degradation using information supplemented by their neighbors without centralized information or processing. Our robots are also not expected to remain in constant communication with each other. Further, we expand our MCP algorithm to correct their assumed accuracy such that environmental estimates (from the collective perception problem) thereafter become more accurate.
\section{Methodology}
\label{sec:methodology}

\subsection{Minimalistic Collective Perception (MCP)}
\label{sec:collective_perception}
Given an environment with a floor patterned with black and white tiles, we pose the problem of collective perception as estimating the black tile fill ratio $f$. The swarm of robots tasked to perform the estimation can do so using their imperfect ground sensor which records the correct tile color with a true accuracy of $b$ (for black tiles) and $w$ (for white tiles). These accuracies represent the probability of observing the correct respective colors, \textit{i.e.,} $b, w \in [0, 1]$. Then, at any time step $k$, the probabilities of making an observation $z^k$ are
\begin{align}
    p(z^k = 1 = \text{observe black}) & = b f + (1 - w)(1 - f),\label{eq:prob_obs_black} \\
    p(z^k = 0 = \text{observe white}) & = (1 - b)f + w(1 - f),\label{eq:prob_obs_white}
\end{align}
derived from the law of total probability.

\begin{figure}[t]
    \centering
    \def\svgwidth{0.99\columnwidth}
    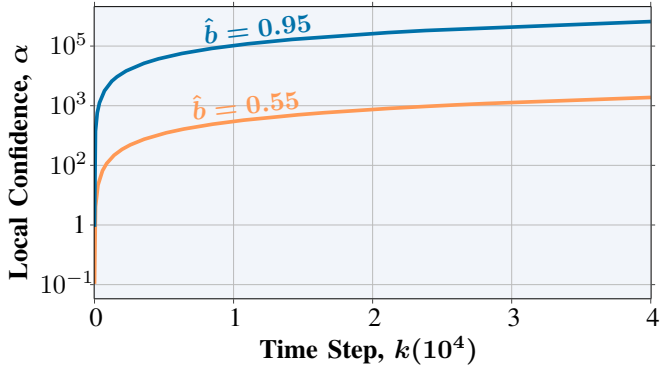
    \caption{With a constant $\sfrac{n}{t}$, a higher $\hat{b}$ robot has a local confidence value $>100\times$ greater than that of a lower $\hat{b}$ robot (\Cref{eq:local_confidence}). Here we assume $\hat{w} = \hat{b}$.}
    \label{fig:no_filter_stsd_alpha_comparison}
\end{figure}

In our previous work~\cite{chinMinimalisticCollectivePerception2023}, the robots using the MCP algorithm are aware of their true accuracies ($b$, $w$). These values become their assumed accuracies ($\hat{b} \coloneq b$, $\hat{w} \coloneq w$) which are used in estimating $f$. In what follows, we outline the main equations from that work to motivate the design of our degradation filter. Each robot diffuses through the environment and makes probabilistic observations of tile colors. Using these observations, the robots independently compute an estimate of the fill ratio, defined as the local estimate, by employing the Binomial likelihood model. This likelihood model stems from the probability (at the $k$\textsuperscript{th} time step) of robot $i$ observing $n^k_i \in \mathbb{Z}^+$ black tiles over $t^k \in \mathbb{Z}^+$ observations:
\begin{align*}
    p\bigg( \sum_{l=1}^{t^k} z^l_i = n^k_i \bigg) = {t^k \choose n^k_i} \big( & \hat{b}^k_{i} f + (1 - \hat{w}^k_{i})(1 - f)\big)^{n^k_i} \cdot    \\
    \big(1 - (                                                                & \hat{b}^k_{i} f + (1 - \hat{w}^k_{i})(1 - f)) \big)^{t^k - n^k_i}.
\end{align*}
Hereafter we consider a single observation is made at each time step, \textit{i.e.,} $t^k = k$, and drop the usage of index $k$ to declutter the notation. Then, robot $i$'s local estimate, $\hat{x}_i$ is found by solving for the maximum likelihood estimator (MLE) of $f$
\begin{equation*}
    \frac{\partial}{\partial f} \ln p\bigg( \sum_{l=1}^{t} z^l = n \mathrel{\bigg|} f \bigg) = 0
\end{equation*}
which yields:
\begin{equation}\label{eq:local_estimate}
    f_{MLE} = \hat{x}_i =
    \begin{cases}
        0                                                                          & \text{if } \dfrac{\nicefrac{n_i}{t} + \hat{w}_{i} - 1}{\hat{b}_{i} + \hat{w}_{i} - 1} \leq 0, \\[10pt]
        1                                                                          & \text{if } \dfrac{\nicefrac{n_i}{t} + \hat{w}_{i} - 1}{\hat{b}_{i} + \hat{w}_{i} - 1} \geq 1, \\[10pt]
        \dfrac{\nicefrac{n_i}{t} + \hat{w}_{i} - 1}{\hat{b}_{i} + \hat{w}_{i} - 1} & \text{otherwise}.
    \end{cases}
\end{equation}
Robot $i$ also gauges its confidence $\alpha_i$ on $\hat{x}_i$ by the fact that the variance lower bound of the MLE is the inverse Fisher information $\mathcal{I}(f)^{-1}$:
\begin{equation*}
    \alpha_i = \mathcal{I}(f) = -\mathbb{E} \left[ \frac{\partial^2}{\partial f^2} \ln{p\bigg( \sum_{l}^{t} z^l = n \bigg)} \mathrel{\bigg|} f \right].
\end{equation*}
The exact form of the local confidence, where $d \coloneq (\hat{b}_{i} + \hat{w}_{i} - 1)^2$, is
\begin{equation}\label{eq:local_confidence}
    \alpha_i =
    \begin{cases}
        \dfrac{d \big( t\hat{w}_{i}^2 - (2\hat{w}_{i} - 1)(t - n_i) \big)}{\hat{w}_{i}^2(\hat{w}_{i}-1)^2} & \text{if } \hat{x}_i = 0, \\[10pt]
        \dfrac{d \big( t\hat{b}_{i}^2 - (2\hat{b}_{i} - 1)n_i \big)}{\hat{b}_{i}^2 (\hat{b}_{i} - 1)^2}    & \text{if } \hat{x}_i = 1, \\[10pt]
        \dfrac{dt^3}{n_i(t - n_i)}                                                                         & \text{otherwise}.
    \end{cases}
\end{equation}
For a fixed $n_i$ and $t$, the confidence increases with the assumed sensor accuracies, \textit{i.e.,} as $\hat{b}_{i}$ and $\hat{w}_{i}$ increase (\Cref{fig:no_filter_stsd_alpha_comparison}).

Next, robots communicate their local values $(\hat{x}, \alpha)$ with their neighbors. From the perspective of robot $i$, these values are accumulated into a social estimate $\bar{x}_i$ and social confidence $\beta_i$:
\begin{align}
    \bar{x}_i & = \frac{1}{\sum_{j \in \mathcal{N}_i} \alpha_j} \sum_{j \in \mathcal{N}_i} \alpha_j \hat{x}_j, \label{eq:social_estimate} \\
    \beta_i   & = \sum_{j \in \mathcal{N}_i} \alpha_j. \label{eq:social_confidence}
\end{align}
where $\mathcal{N}_i$ is the set of robot $i$'s neighbors.

As the last step, robot $i$ combines its local values $(\hat{x}_i, \alpha_i)$ with its social values $(\bar{x}_i, \beta_i)$ into an informed estimate $x_i$:
\begin{align}\label{eq:informed_estimate}
    x_i & = \frac{\alpha_i \hat{x}_i + \beta_i \bar{x}_i}{\alpha_i + \beta_i} = \frac{\alpha_i \hat{x}_i + \sum_{j \in \mathcal{N}_i} \alpha_j \hat{x}_j}{\alpha_i + \sum_{j \in \mathcal{N}_i} \alpha_j}
\end{align}
which is a weighted average of local estimates. This estimate represents the robot's best guess of the environment fill ratio $f$ at step $k$, and the process repeats.

\subsection{Adaptive Sensor Degradation Filter (ASDF)}
\label{sec:sensor_degradation_filter}

\subsubsection{Sensor Accuracy Update}
\label{sec:sensor_degradation_filter_update}
The MCP algorithm described thus far relies on the assumption that the robots know exactly how inaccurate their sensors are,~\textit{i.e.,} the true accuracies $b$ and $w$ are perfectly known to each robot. In the calibration problem, we remove this assumption: the robots' have assumed sensor accuracies $\hat{b}$ and $\hat{w}$ which may not be equivalent to the true accuracies. The calibration scheme thus involves a continuous update to the assumed accuracies such that each robot's informed estimate $x$ remain accurate.

We do this akin to estimating robot $i$'s local estimate, $\hat{x}_i$, of the environment fill ratio $f$. With the black-and-white environment setup, we let $w_i = b_i$---robot $i$'s ability to identify a white tile is equivalent to its ability to identify the opposite color---and flip \Cref{eq:local_estimate} to solve for the assumed accuracy. Doing so allows us to leverage the MLE's optimality, albeit the outcome with the highest likelihood now depends on both $f_{MLE}$ and $\hat{b}_{i}$ being optimal. Solving for the assumed accuracy results in the sensor degradation filter:
\begin{align}\label{eq:assumed_accuracy_update}
    \hat{b}_{i} =
    \begin{cases}
        0                                                         & \text{if } \dfrac{\nicefrac{n_i}{t} + \bar{x}_i - 1}{2\bar{x}_i - 1} \leq 0, \\[10pt]
        1                                                         & \text{if } \dfrac{\nicefrac{n_i}{t} + \bar{x}_i - 1}{2\bar{x}_i - 1} \geq 1, \\[10pt]
        \dfrac{\nicefrac{n_i}{t} + \bar{x}_i - 1}{2\bar{x}_i - 1} & \text{otherwise,}
    \end{cases}
\end{align}
where we use the social estimate $\bar{x}_i$ in place of the unknown fill ratio $f$. In brief, robot $i$ uses its neighbors' local estimate of $f$ as a reference to adjust its understanding of the sensor accuracy $\hat{b}_{i}$. The robots with an incorrectly assumed sensor accuracy are deemed to have a \emph{flawed assumed accuracy}.

\subsubsection{Adaptive Filter Activation}
\label{sec:sensor_degradation_filter_activation}
Given a constant assumed accuracy update rate, there may be situations where a robot may be better served \emph{not} to activate the filter. An instance that we observed in preliminary experiments (not shown in this work) is one where a robot that starts out with a correct assumed accuracy ($\hat{b}_{i} = b_i$) ends up with an incorrect one ($\hat{b}_{i} \neq b_i$) due to unnecessary filter updates. This happens because the environment's stochasticity implies that each robot's local estimate $\hat{x}$ are unlikely to be the same at any moment, especially in early observations. These minor deviations manifest in a robot's social estimate $\bar{x}_i$ that in turn perturbs its estimated assumed accuracy $\hat{b}_{i}$ in \Cref{eq:assumed_accuracy_update} away from the initially correct value. A feedback loop is thus created: robot $i$'s incorrect $\hat{b}_{i}$ leads to an inaccurate $\hat{x}_i$ that is communicated to neighboring robots $j$, affecting their assumed accuracy $\hat{b}_{j}$. Eventually, with most of the robots having an incorrect $\hat{b}$, the swarm settles on an inaccurate informed estimate $x$.

For this reason, we extended the algorithm to self-identify filter activation conditions. A glaring challenge in devising such a scheme is the lack of information regarding which of the neighbors to trust, or when to trust them. Our approach addresses this by way of a hypothesis test that evaluates whether robot $i$'s local estimate $\hat{x}_i$ is similar to its neighbors'. If the estimates are too different, robot $i$ updates its assumed accuracy based on \Cref{eq:assumed_accuracy_update}. This effectively promotes conformity in $\hat{x}$ within the swarm, which improves convergence in $x$---though, as we observe in the simulated results (\Cref{sec:simulated_experiments}), the swarm's consensus may not always be accurate.

From the viewpoint of a robot $i$, we begin the derivation by assuming that our neighbors' local estimates follow a Gaussian distribution with an unknown mean, $\mu$ and variance, $\sigma^2$: $\hat{x}_j \overset{i.i.d.}{\sim} \mathcal{N}(\mu, \sigma^2)$. This assumption stems from the asymptotic normality of MLEs, which we show in \Cref{appdx:asymp_norm}. Our test is set up with the null hypothesis stating that our local estimate $\hat{x}_i$ is similar to the average neighbor estimate (among $m_i \geq 2$ neighbors):
\begin{align*}
    H_0: \mu = \hat{x}_i, \qquad H_1: \mu \neq \hat{x}_i.
\end{align*}
We then obtain the following rejection region $R_i$ for robot $i$ by the procedure of defining a likelihood ratio test, based on the Gaussian model assumption:
\begin{align*}
    R_i & = \bigg\{ \hat{x}_j: \frac{|\bar{X}_i - \hat{x}_i|}{\nicefrac{S_i}{\sqrt{m_i}}} > c \bigg\}
\end{align*}
where the sample average and sample variance are
\begin{align*}
    \bar{X}_i = \frac{\sum_j^{m_i}{\hat{x}_j}}{m_i} \text{ and } S_i^2 = \frac{\sum_j^{m_i}{(\hat{x}_j - \bar{X}_i)^2}}{m_i - 1}
\end{align*}
respectively and $c \geq 0$ is the rejection threshold. Note that $c$ is found using the power function and a user-specified error probability. The complete derivation of the test is detailed in \Cref{appdx:lrt_derivation}.

Using $R_i$, we define the statistical power function $\gamma(\mu)$, \textit{i.e.,} the probability of rejecting the null hypothesis, based on a user-specified type II error probability $\omega$. ($\omega$ is equivalent to the false negative rate, the probability that the null hypothesis is mistakenly \textit{not rejected}; in our case, $\omega$ is the rate that robot $i$ incorrectly `thinking' that its local estimate $\hat{x}_i$ is similar to its neighbors'.) The definition expands to a familiar result where our test statistic $T$ follows a $t_{m_i - 1}$ distribution:
\begin{align} \label{eq:power_func}
    \begin{split}
        \gamma(\mu) & = p(\hat{x}_j \in R_i)                                                                                                                                                     \\
                    & = p\bigg( \frac{|\bar{X}_i - \hat{x}_i|}{\nicefrac{S_i}{\sqrt{m_i}}} > c \bigg)                                                                                            \\
                    & = 2 p\bigg( \frac{\bar{X}_i - \hat{x}_i}{\nicefrac{S_i}{\sqrt{m_i}}} > c \bigg)                                                                                            \\
                    & = 2 p\bigg( \frac{\bar{X}_i - \mu}{\nicefrac{S_i}{\sqrt{m_i}}} - \frac{\hat{x}_i - \mu}{\nicefrac{S_i}{\sqrt{m_i}}} > c \bigg)                                             \\
                    & = 2 p\bigg( T > c + \frac{\hat{x}_i - \mu}{\nicefrac{S_i}{\sqrt{m_i}}} \bigg) \qquad \qquad \bigg\{ T \coloneq \frac{\bar{X}_i - \mu}{\nicefrac{S_i}{\sqrt{m_i}}} \bigg \} \\
                    & = 1 - \omega.
    \end{split}
\end{align}

To solve for $c$, we consider the power function at one standard deviation $S$ away from the null hypothesis: $\gamma(\hat{x}_i + S_i) = 1 - \omega$. The solution is thus
\begin{align} \label{eq:power_func_k_solution}
    \gamma(\hat{x}_i + S_i) & = 2 p(T > c + \frac{\hat{x}_i - \hat{x}_i - S_i}{\nicefrac{S_i}{\sqrt{m_i}}}) \notag \\
                            & = 2p(T > c - \sqrt{m_i}) \notag                                                      \\
                            & = 1 - \omega, \notag                                                                 \\
    \Rightarrow c           & = t_{m_i - 1, \frac{1 - \omega}{2}} + \sqrt{m_i}
\end{align}
where $t_{m_i - 1, \frac{1 - \omega}{2}}$ is the $t$-score such that $p(T > t_{m_i - 1, \frac{1 - \omega}{2}}) = (1 - \omega)/2$ and the complete definition of our rejection region becomes
\begin{align}\label{eq:rejection_region}
    R = \bigg\{ \hat{x}_j: \frac{|\bar{X}_i - \hat{x}_i|}{\nicefrac{S_i}{\sqrt{m_i}}} > c = t_{m_i - 1, \frac{1-\omega}{2}} + \sqrt{m_i} \bigg\}.
\end{align}

Therefore, as robot $i$, we reject that our local estimate is similar to our neighbors' and update our assumed sensor accuracy whenever the following condition is satisfied:
\begin{align}\label{eq:hypothesis_test}
    |\bar{X}_i - \hat{x}_i| >  \bigg( \frac{t_{m_i - 1, \frac{1-\omega}{2}}}{\sqrt{m_i}} + 1 \bigg) S_i.
\end{align}
Due to the usage of a sample mean of neighboring estimates, we expect the adaptive activation scheme to perform best when flawed robots are in the minority. Note that this method needs only the specification of a desired type II error probability ($\omega$) and is adaptive to the number of neighbors encountered ($m_i$), as well as the mean ($\bar{X}_i$) and variance (${S_i^2}$) of their estimates.

Since no additional information is exchanged between robots for the ASDF, the computational and bandwidth complexity is $O(1)$. Space complexity is $O(q)$ due to the storage of the $t$-test table with $q$ elements. The ASDF runs as part of the MCP algorithm, executing after the informed estimation step to update the accuracy $\hat{b}$ for the next iteration. The complete algorithm is shown in Algorithm \ref{alg:sensor_degradation_filter}.

\begin{algorithm}[t]
    \footnotesize
    \hsize=\columnwidth
    \caption{MCP-ASDF executed by an individual robot}\label{alg:sensor_degradation_filter}
    \begin{algorithmic}
        \Require $0 < \hat{b} = \hat{w} < 1$ \Comment{sensor accuracies}
        \Ensure $x$ \Comment{informed estimate}
        \State $t \gets 0$ \Comment{total tiles observed}
        \State $n \gets 0$ \Comment{total black tiles observed}
        \While{\Not{} $\Call{Done}{}$}
        \State $\Call{Move}{}$
        \State $z \gets \Call{ObserveEnvironment}{}$
        \State $t \gets t + 1$
        \If{$z$ = Black}
        \State $n \gets n + 1$
        \EndIf
        \State $(\hat{x}, \alpha) \gets \Call{LocalEstimation}{t, n, \hat{b}, \hat{w}}$ \Comment{eqs.~\ref{eq:local_estimate}, \ref{eq:local_confidence} with $\hat{w} = \hat{b}$}
        \State $\mathbf{H} \gets \Call{DetectNeighbors}{}$ \Comment{generate list of neighbors}
        \If{$\mathbf{H} \neq$ empty}
        \State $\mathbf{\hat{x}_H}, \mathbf{\alpha_H} \gets \Call{Receive}{\mathbf{H}}$
        \State $\Call{Broadcast}{\mathbf{H}, \hat{x}, \alpha}$
        \State $(\bar{x}, \beta) \gets \Call{Social Estimation}{\mathbf{\hat{x}_H}, \mathbf{\alpha_H}}$ \Comment{eqs.~\ref{eq:social_estimate}, \ref{eq:social_confidence}}
        \EndIf
        \State $x \gets \Call{Informed Estimation}{\hat{x}, \alpha, \bar{x}, \beta}$ \Comment{eq.~\ref{eq:informed_estimate}}
        \If{\Call{Activate}{$\mathbf{\hat{x}_H}$}} \Comment{eq.~\ref{eq:hypothesis_test}}
        \State $\hat{b} \gets \Call{Update Accuracy}{\mathbf{\hat{x}_H}}$ \Comment{eq.~\ref{eq:assumed_accuracy_update}}
        \EndIf
        \EndWhile
    \end{algorithmic}
\end{algorithm}
\section{Simulated Experiments}
\label{sec:simulated_experiments}

\subsection{General Setup and Metrics}
\label{sec:simulated_setup_metrics}

\paragraph{Setup}
Our simulated experiments are run with a group of $N = 20$ robots using the ARGoS multi-robot, physics-based simulator~\cite{pinciroliARGoSModularParallel2012}. Each experiment is replicated with a different random seed for $M = 30$ times, each trial lasting for $k_{max} = 40,000$ time steps. The experiments are conducted in a $L \times L$ square arena and populated with black and white tiles, as parametrized by the target fill ratio $f$. The robots diffuse across the arena and are modeled after the Khepera IV~\cite{soaresKheperaIVMobile2016}. These robots have differential drive and proximity sensors as well as a communication range of 5 body lengths at $r = \unit[0.7]{m}$. They travel at a maximum forward speed of $\unit[0.14]{ms^{-1}}$; this is the speed required to traverse along the diagonal of a single tile in one time step. In each of the experiments, all the robots have the same true accuracy $b$ in observing tile colors, but not all of them have their assumed accuracy $\hat{b}$ initialized to that. A proportion $P$ of the robots have a flawed assumed accuracy $\hat{b}$ initially. As the experiment progresses, any robot running the ASDF updates its assumed accuracy, regardless of whether $\hat{b}$ was correct to begin with. We use a type II error probability of $\omega = 0.05$ to parametrize the adaptive activation mechanism.

To contextualize the physics-based experimental results---where communication between robots happens via a dynamic network topology---we include simulated results of 10 robots in a fully connected network topology. As it represents the ideal communication network---assuming an unconstrained bandwidth---full connectivity provides maximum information exchange, thus the estimation performance under such conditions serves as an upper limit reference. Experiments for the fully connected swarm are executed using a custom Python simulator similar to our previous work~\cite{chinMinimalisticCollectivePerception2023}. Each robot has direct communication channels to every other robot, but it \emph{does not} undergo any physical motion. The intended diffusion motion is instead simulated through provisioning a sequence of tile color observations that follow a Bernoulli distribution (parametrized by $f$). Alternatively, these robots can be thought of as nodes in a static, fully connected graph that each processes a sequence of Bernoulli trials (observations $z$); after each time step, they pass messages between one another (communicated local values $\hat{x}$ and $\alpha$). Our motivation behind such abstraction reflects the emphasis on creating an ideal communication setup. The code used in running both the dynamic and fully connected topology experiments is available in our project repository.\footnote{\url{https://github.com/NESTLab/sensor-degradation-filter}}

Over different experiments, we vary the following parameters for a comprehensive investigation of the filter performance:
\begin{itemize}
    \item flawed robot proportions, $P = \{ 0, 10, 30, 50, 100 \}\%$;
    \item nominal filtering periods, $\tau = \{ 1000, 2000, 4000 \}$;
    \item true accuracies, $b = \{ 0.55, 0.95 \}$;
    \item flawed assumed accuracies, $\hat{b} = \{ 0.55, 0.75, 0.95 \}$;
    \item target fill ratios, $f = \{\text{ambiguous, unambiguous}\} = \{ 0.55, 0.95 \}$.
\end{itemize}
For the dynamic topology experiments, we controlled the swarm density, defined as $D = N \pi r^2 / L^2$, to be 1. Note that the swarm density parameter does not apply to the fully connected robots as information exchange happens irrespective of communication range.

\paragraph{Metrics}
Our main metrics of interest are the \emph{convergence speed} and \emph{accuracy} of the informed estimates $x$ generated by the swarm. Convergence speed is evaluated using the time step $K$ at which an informed estimate `settles' to within a specified threshold $\delta$ ($0.01$ in this study) until the experiment ends: $|x^K - x^k| < \delta$, $k \geq K$. The greater $K$ is the lower the convergence speed. Accuracy is calculated with the absolute error $e$ of the informed estimate at $K$: $e = |x^K - f|$. A high accuracy is attributed to a low $e$.

Furthermore, we devised a joint metric that combines both an estimate's convergence time and absolute error into a single score ranging between 0 and 100 to facilitate the discussion of the algorithm's performance. First, for each robot $i$'s informed estimate, we compute its convergence speed and accuracy scores respectively as
\begin{align}\label{eq:separate_score}
    h_{K,i} = 100 \bigg( \frac{K_{max} - K}{K_{max}} \bigg), \quad h_{e,i} = 100 \bigg( \frac{e_{max} - e}{e_{max}} \bigg)
\end{align}
where $K_{max}$ and $e_{max}$ are the maximum values for the convergence time and absolute error, respectively. In the discussion that follows, $K_{max}$ and $e_{max}$ are found based on the particular experimental configuration under analysis. This allows us to isolate the evaluation of a set of experiments without having it tampered with by another that may be distributed differently. Setting a fixed value for $K_{max}$ and/or $e_{max}$ would instead confine both sets to the same standard, which can obscure performance distinctions under a unified analysis.

Aside from convergence speed and accuracy, the performance score also considers the spread of all robots' estimates within each trial to account for consensus. To that end, the \emph{convergence and accuracy scores for a single trial} are each a sample mean of respective individual scores across $N$ robots scaled by the interquartile range $IQR$:
\begin{align}\label{eq:consensus_scaling}
    \begin{split}
        h_K & = \frac{\sum_i^N h_{K,i}}{N} \exp \bigg\{ \frac{-IQR_K}{K_{max}} \bigg\}, \\
        h_e & = \frac{\sum_i^N h_{e,i}}{N} \exp \bigg\{ \frac{-IQR_e}{e_{max}} \bigg\}.
    \end{split}
\end{align}
Finally, the \emph{combined score of a swarm for a single trial}, $H$ is just the average of the scaled scores:
\begin{align}\label{eq:overall_score}
    H & = \frac{1}{2} (h_K + h_e).
\end{align}
This is the metric used in our evaluation of the filter's effectiveness under various operating conditions and parametrizations.

\subsection{Factors Affecting Behaviors of Individual Robots}
Before diving into the experimental results, we briefly outline several factors responsible for a robot's estimation behavior. They only account for behaviors of an \emph{individual robot} (a swarm's emergent performance will be analyzed specifically in \Cref{sec:simulated_baseline_performance,sec:simulated_asdf_perf_false,sec:simulated_asdf_perf_true}). We will refer to the factors as we examine the estimation behaviors in the discussion that follows. In some cases, the estimation is affected by a combination of the factors below, yielding an outcome that can be unexpected.

\begin{factor}{Environment}{Specific combinations of fill ratio and true accuracy may be easier for estimation than others} \label{factor:specific_f_bt_combo}
    Revisiting \Cref{eq:prob_obs_black,eq:local_estimate} indicates to us that not all pairings of environment ambiguity ($f$) and true sensor accuracies ($b$) are equal in challenging a robot's estimation capabilities. As \Cref{tab:prob_obs_black_and_local_est} shows, the local estimates under certain conditions are extremely close to the environment fill ratio, which is likely to \emph{improve estimation accuracy and convergence}. This is especially relevant for the discussion on no-filtering robots in \Cref{sec:simulated_baseline_performance}, but also applicable to other cases.
\end{factor}

\begin{factor}{Volatility}{Lower assumed accuracies increase local estimate volatility} \label{factor:lower_ba_volatile}
    This stems from the assumed accuracy's role in computing the local estimate (\Cref{eq:local_estimate}). Suppose that there are two robots with assumed accuracies $\hat{b} = 0.55$ and $\hat{b} = 0.95$. With a similar average number of black tiles observed $\sfrac{n}{t}$, the rate of change in $\hat{x}$ is higher for the robot with a lower $\hat{b}$. As a consequence, the robot with a more volatile local estimate would suffer from a \emph{slower informed estimate convergence} with minimal neighbor contact.
\end{factor}

\begin{factor}{Period}{Lower filtering periods are better at mitigating the performance decline} \label{factor:lower_t_better}
    With shorter nominal filtering periods (`nominal' because filter activation is adaptive), the estimation performance performs better than longer periods overall. This is because as the duration between filter updates grows, a robot's estimation performance gradually reverts into the no-filtering case.
\end{factor}

\begin{factor}{Neighbor}{Higher assumed accuracy neighbors have more influence in informed estimation} \label{factor:higher_ba_overrides}
    This factor is caused by the computation of $x$ as a weighted average in \Cref{eq:informed_estimate}. As \Cref{fig:no_filter_stsd_alpha_comparison} depicts, the weights $\alpha$ are an increasing function of the assumed accuracy $\hat{b}$, which means that the local estimates from higher $\hat{b}$ neighbors have an oversized influence. This can either work \emph{for} or \emph{against a robot's estimation performance}, depending on whether the neighbors' higher $\hat{b}$ values are warranted. The effect is similar to those in works where robots communicate opinions~\cite{talamaliWhenLessMore2021,crosscombeImpactNetworkConnectivity2022}, as the more influential transmitter robots affect the behavior of recipient robots.
\end{factor}

\begin{table}[!t]
    \centering
    \caption{Black tile observation probability $p(z^k = 1)$ (\Cref{eq:prob_obs_black}) and local estimate $\hat{x}^k_i$ (\Cref{eq:local_estimate}) of a robot $i$ based on various environment fill ratios $f$ and true sensor accuracies $b_i$. Thanks to the \Cref{factor:specific_f_bt_combo}, local estimation is easier when $f = b_i$, regardless of $\hat{b}^k_i$. Note that the index $k$ and $i$ have been omitted for brevity.}
    \label{tab:prob_obs_black_and_local_est}
    \def\arraystretch{2}%  1 is the default, change whatever you need
    \begin{tabular}{|m{0.03\linewidth}<{\centering}||m{0.405\linewidth}<{\centering}|m{0.405\linewidth}<{\centering}|}
        % \toprule
        \hline
         & $b = 0.55$ & $b = 0.95$ \\
        % \cmidrule(rl){2-2} \cmidrule(rl){3-3}
        % \hline
        \hhline{|=#=|=|}
        \rotatebox{90}{$f = 0.55$}
         & {
                \begin{gather*}
                    p(z = 1) = \frac{n}{t} = 0.505 \\
                    \hat{x} =
                    \begin{cases}
                        0.550, & \hat{b} = b       \\
                        0.510, & \hat{b} = b + 0.2 \\
                        0.506, & \hat{b} = b + 0.4
                    \end{cases}
                \end{gather*}
            }
         & {
                \begin{gather*}
                    p(z = 1) = \frac{n}{t} = 0.545 \\
                    \hat{x} =
                    \begin{cases}
                        0.550, & \hat{b} = b       \\
                        0.590, & \hat{b} = b - 0.2 \\
                        0.950, & \hat{b} = b - 0.4
                    \end{cases}
                \end{gather*}
        }                          \\
        % \cmidrule(rl){2-2} \cmidrule(rl){3-3}
        \hline
        \rotatebox{90}{$f = 0.95$}
         & {
                \begin{gather*}
                    p(z = 1) = \frac{n}{t} = 0.545 \\
                    \hat{x} =
                    \begin{cases}
                        0.950, & \hat{b} = b       \\
                        0.590, & \hat{b} = b + 0.2 \\
                        0.550, & \hat{b} = b + 0.4
                    \end{cases}
                \end{gather*}
            }
         & {
                \begin{gather*}
                    p(z = 1) = \frac{n}{t} = 0.905 \\
                    \hat{x} =
                    \begin{cases}
                        0.950, & \hat{b} = b       \\
                        1.000, & \hat{b} = b - 0.2 \\
                        1.000, & \hat{b} = b - 0.4
                    \end{cases}
                \end{gather*}
        }                          \\
        % \bottomrule
        \hline
    \end{tabular}
\end{table}

\subsection{Baseline Performance}
\label{sec:simulated_baseline_performance}

% Combined score boxplot: BRAVO filter P = 0 to 100%
\begin{figure*}[!t]
    \centering
    \begin{subfigure}{\textwidth}
        \centering
        \def\svgwidth{\textwidth}
        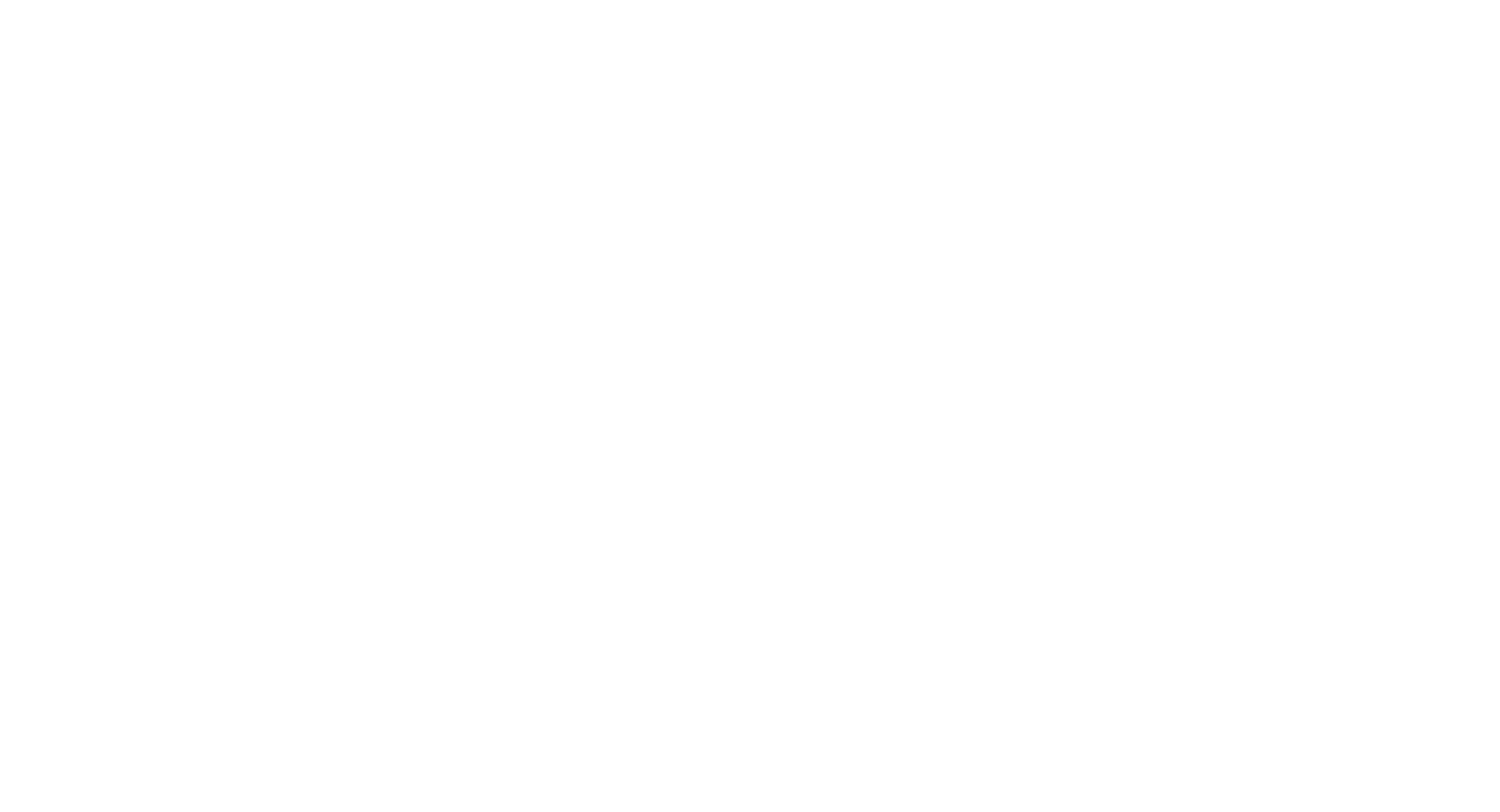
        \caption{Scores when \emph{only flawed robots} execute the ASDF. Filtering at $\tau = 1000$ (orange), overconfident swarms are just as effective as a flawless one, though such robustness extends only to swarms with at most $10\%$ underconfident robots.}
        \label{fig:BRAVOFalse_dtsd_den1_noP100}
    \end{subfigure}
    \par \bigskip
    \begin{subfigure}{\textwidth}
        \centering
        \def\svgwidth{\textwidth}
        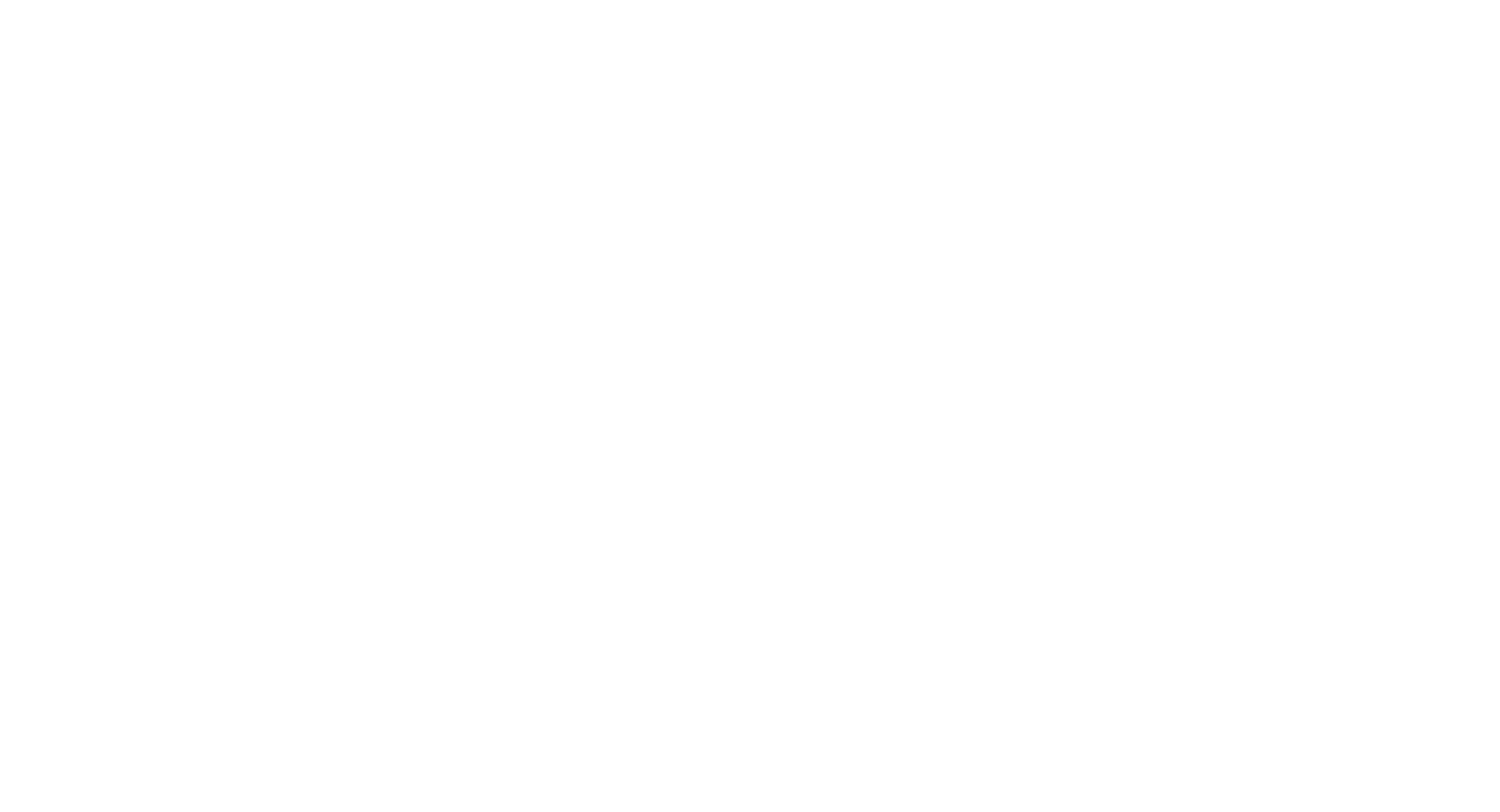
        \caption{Scores when \emph{all robots} execute the ASDF. Except for a swarm of $P = 50\%$ underconfident robots filtering at $\tau \geq 2000$ at $f = 0.95$ (bottom right), the filter successfully improves estimation over no-filtering robots.}
        \label{fig:BRAVOTrue_dtsd_den1}
    \end{subfigure}
    \caption{Comparison of informed estimation performance across various nominal filtering periods $\tau$ (box colors), where each box represents 30 trial scores, $H$, from \Cref{eq:overall_score}. Higher $H$ is preferred as it indicates better performance. Scores of experiments with different flawed assumed accuracies $\hat{b}$ are shown (secondary $x$-axis) as deviations from $b$, plotted for each $P$ (primary $x$-axis). At $P = 0\%$, all robots have correct assumed accuracies, hence the absence of a secondary $x$-axis label. The \textit{No Filtering} data is identical in both (a) and (b). Both sets of scores are computed with $K_{max} = 40000$, $e_{max} = 0.45$.}
    \label{fig:BRAVO_dtsd_den1}
\end{figure*}

\paragraph{Dynamic robots}
First, we examine the informed estimation performance of the swarm \emph{without any filtering}, which will be used to frame filtering performances in the later subsections. This means that the swarm performs MCP as outlined in \Cref{sec:collective_perception} with no awareness that their assumed accuracy $\hat{b}$ may be flawed. The results are shown in \Cref{fig:BRAVO_dtsd_den1}.

\begin{figure*}[!t]
    \centering
    \begin{subfigure}{\textwidth}
        \centering
        \def\svgwidth{0.65\textwidth}
        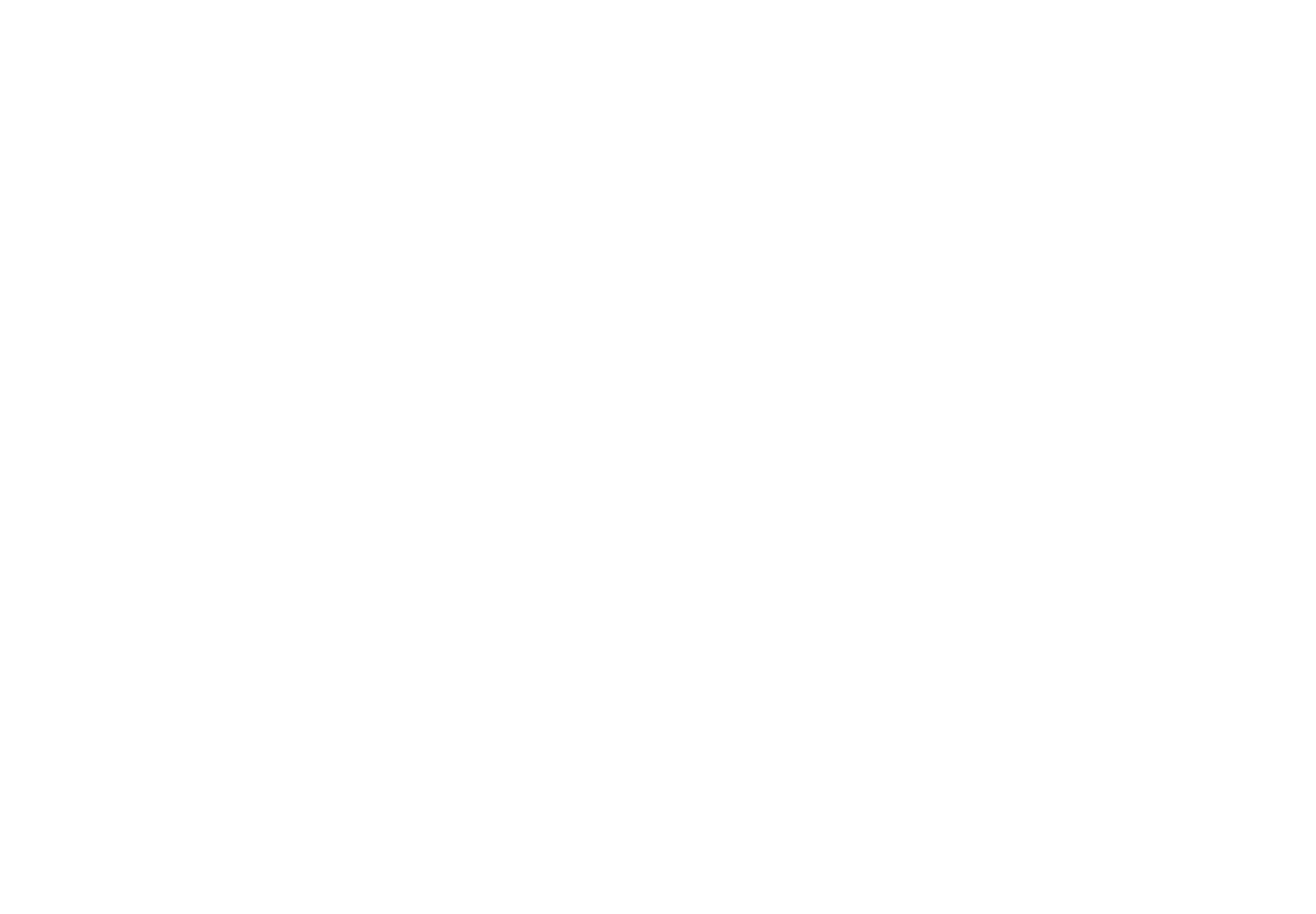
        \caption{Scores of the no-filtering experiment. In general, swarms with overconfident robots (blue dots) trade accuracy for convergence speed with the increase of flawed robots, though more so in an unambiguous environment ($f = 0.95$) than in an ambiguous one ($f = 0.55$). Swarms with underconfident robots (red/orange dots), however, may thrive with more flawed robots depending on environmental conditions. Note that in the bottom right panel data points for both underconfident cases are atop each other.}
        \label{fig:no_filtering_sep_dtsd_den1}
    \end{subfigure}
    \par \bigskip
    \begin{subfigure}{\textwidth}
        \centering
        \def\svgwidth{\textwidth}
        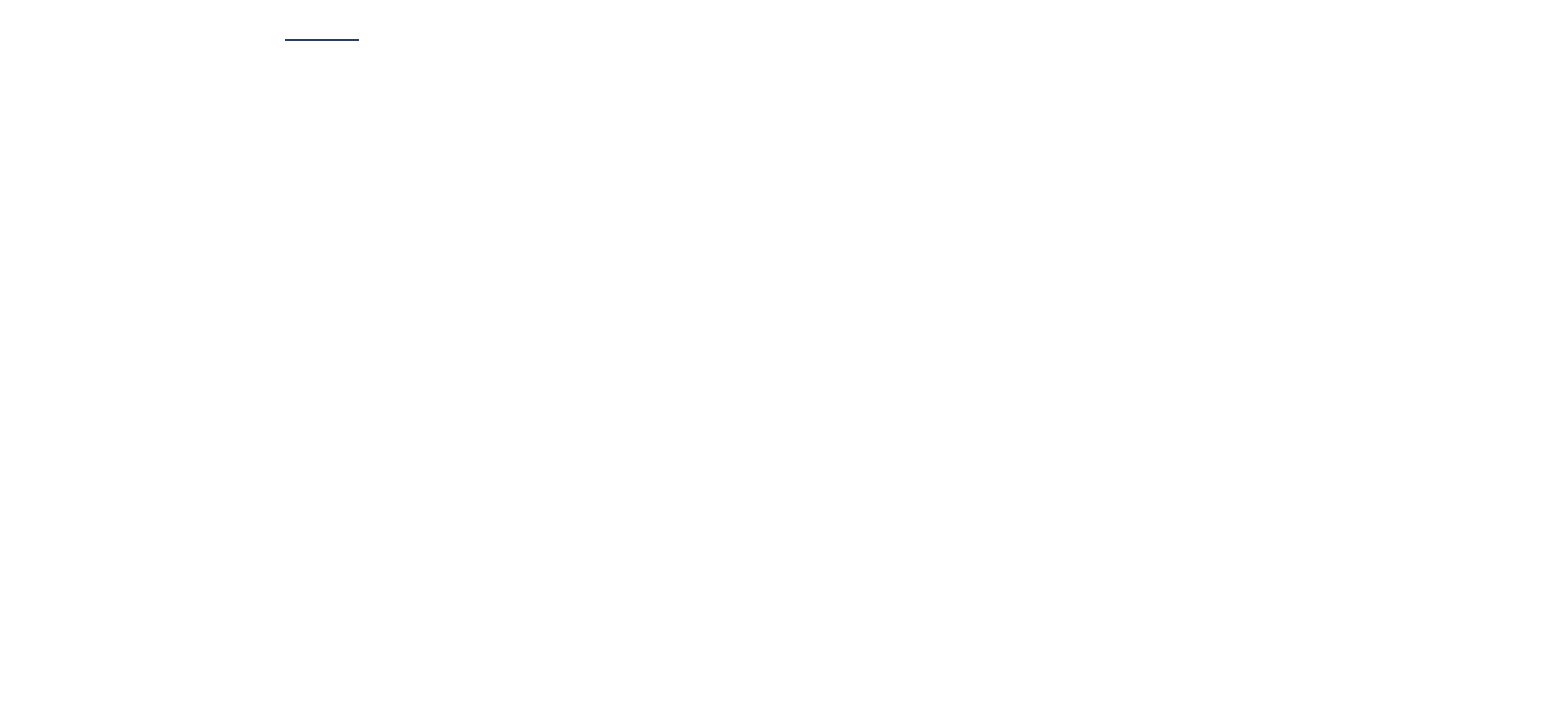
        \caption{Scores of the ASDF experiments. For partially flawed swarms, both filters exhibit similar trends whereby estimation accuracy is excellent generally; though, due to the \Cref{factor:higher_ba_overrides} overconfident \ASDFp{} robots (blue dots) are susceptible to adopting incorrect assumed accuracies that degrade accuracy scores.}
        \label{fig:BRAVO_sep_dtsd_den1}
    \end{subfigure}
    \caption{Comparison of informed estimation performance in terms of convergence and accuracy, where each point represents the $h_K$ and $h_e$ scores (from \Cref{eq:consensus_scaling}) for a single trial. Higher $h_K$ (rightward) and $h_e$ (upward) scores are preferred as they indicate better performance. Both sets of scores are computed with $K_{max} = 40000$, $e_{max} = 0.45$.}
    \label{fig:sep_dtsd_den1}
\end{figure*}

Between $P = 0\%$ to $50\%$ (\Cref{fig:BRAVOFalse_dtsd_den1_noP100}), a swarm's ability to provide accurate and timely estimates deteriorates with an increasing percentage of flawed robots. An exception to this is when $f = 0.55$, $b = 0.55$ (top left), where we observe performance improving with more overconfident robots (\textit{i.e.,} having overestimated assumed accuracies). In fact, having any number of flawed robots improves performance over a flawless swarm. The reason behind this peculiar outcome is due to the combined effects of the \Cref{factor:specific_f_bt_combo,factor:lower_ba_volatile}. The former enables the swarm to maintain a fairly high estimate accuracy while the latter significantly boosts convergence speed. On top of that, the growing number of overconfident robots implies that the \Cref{factor:higher_ba_overrides} further enhances the speed of converging onto an estimate that is only slightly less accurate. This is revealed in \Cref{fig:no_filtering_sep_dtsd_den1}: informed estimates shift rightward with a slight downward movement as $P$ increases when $b = 0.55$.

% % Separate score scatter plot: No filtering DTSD
% \begin{figure*}[!t]
%     \centering
%     \includegraphics[width=0.85\textwidth]{figures/no_filtering_sep_dtsd_den1.pdf}
%     \caption{Uncoupled informed estimation performance of the no-filtering swarm, where each point represents the $h_K$ and $h_e$ scores (from \Cref{eq:consensus_scaling}) for a single trial, computed with $K_{max} = 40000$, $e_{max} = 0.45$.}
%     \label{fig:no_filtering_sep_dtsd_den1}
% \end{figure*}

% % Separate score scatter plot: No filtering DTSD
% \begin{figure*}[!h]
%     \centering
%     \includegraphics[width=\textwidth]{figures/BRAVOFalse+True_sep_dtsd_den1.pdf}
%     \caption{Comparison of uncoupled informed estimation performance between the \ASDFm{} and \ASDFp{}, where each point represents the $h_K$ and $h_e$ scores (from \Cref{eq:consensus_scaling}) for a single trial, computed with $K_{max} = 40000$, $e_{max} = 0.45$.}
%     \label{fig:BRAVO_sep_dtsd_den1}
% \end{figure*}

% Separate score scatter plot: No filtering DTSD

Interestingly, the estimate performance by an entire swarm of flawed robots ($P = 100\%$) performs \emph{better} than in situations where only some or none of the robots are flawed ($P < 100\%$), for almost all the cases shown in \Cref{fig:BRAVOTrue_dtsd_den1}. In situations where the entire swarm is overconfident, we see the influence of the \Cref{factor:lower_ba_volatile}: the high $\hat{b}$ reduces local estimate volatility, manifesting in a more consistent informed estimate. This is also aided by the \Cref{factor:higher_ba_overrides}: accurate local estimates are weighted less when computing for the informed estimate. The presence of the \Cref{factor:lower_ba_volatile,factor:higher_ba_overrides} are especially evident when comparing two cases in \Cref{tab:prob_obs_black_and_local_est} with seemingly symmetric local estimate errors: $\hat{x} = 0.55$ in an unambiguous environment (top right, $\hat{b} = 0.55 + 0.4$) \textit{v.s.} $\hat{x} = 0.95$ in an ambiguous environment (bottom left, $\hat{b} = 0.95 - 0.4$). Despite both having similar $\sfrac{n}{t}$ values, the entire swarm of flawed, overconfident robots converges faster on an informed estimate than the underconfident ones (\Cref{fig:no_filtering_sep_dtsd_den1}). Notice also that the estimate accuracy is comparable between the two, indicating that the higher combined score is solely a consequence of a superior convergence speed. Conversely, when the entire swarm is underconfident at $f = 0.55$, the \Cref{factor:lower_ba_volatile} dominates: the local estimate is \emph{more} volatile with a lower $\hat{b}$. Nonetheless, the \Cref{factor:specific_f_bt_combo} prevails for the entire swarm of underconfident robots at $f = 0.95$ (\Cref{fig:BRAVOTrue_dtsd_den1}, bottom right) because \Cref{eq:local_estimate} saturates $\hat{x}$ to $1.0$ most of the time (\Cref{tab:prob_obs_black_and_local_est}). Effectively, as can be seen from \Cref{fig:no_filtering_sep_dtsd_den1}, this removes much of the volatility and accelerates convergence, hence the high performance score.

\paragraph{Fully connected robots}
To put the dynamic topology results into perspective, we consider the swarm's performance in a fully connected network topology as a reference, shown in \Cref{fig:BRAVO_stsd_v_dtsd_den1_commsp1}. The performance of fully connected, \emph{overconfident} robots are largely similar (besides the expected positive offset) to those in the dynamic topology experiments: it gradually increases when the environment is ambiguous (top left), while it falls with the rise of flawed robot proportions when the environment is unambiguous (bottom left). In contrast, fully connected, \emph{underconfident} robots outdo their dynamic topology counterparts substantially: the growing number of flawed robots does not weaken informed estimation performance in any way, until when all robots are flawed (which was explained before with dynamic robots).

We attribute this departure in trend to the difference in determinism of information exchange. Each of the fully connected robots has complete knowledge of local estimates, enabling the swarm to act as a multi-node sensor network. That is, the informed estimate computed in \Cref{eq:informed_estimate} is identical for every robot. When these flawed robots are underconfident, their propagated estimates carry less weight as compared to estimates by correct robots, whose confidence value $\alpha$ is higher. As a consequence, local estimates from underconfident robots have a diminishing influence in the informed estimate calculation (such behavior for flawed overconfident robots is shown in \Cref{fig:stsd_corfilt1_timeseries_comparison}). This is effectively an amplified version of the \Cref{factor:higher_ba_overrides}. The benefit of consistently having higher-weighted and correct estimates, however, is absent when robots interact randomly in the dynamic topology setup, which is why we see the stark performance difference.

In a practical setting, where it may be challenging to parametrize the MCP with the correct assumed accuracy, informed estimation of even a partially flawed swarm suffers. The performance decline is especially pronounced for swarms meant for field deployments: with only localized communication available at their disposal, the robots' estimation performance would likely follow results shown in the dynamic topology experiments. This creates an opening that our proposed approach, the ASDF, aims to address. To be concise, we shall refer to the filter where \emph{only flawed robots} run it as \ASDFm{} (\Cref{sec:simulated_asdf_perf_false}) and where \emph{all robots} run it as \ASDFp{} (\Cref{sec:simulated_asdf_perf_true}); otherwise, ASDF is used when referring to both.

\subsection{\ASDFm{} Performance: Flawed Robot Identities are Known}
\label{sec:simulated_asdf_perf_false}
\paragraph{Dynamic robots}
We begin by considering a situation where flawed robots are identified \textit{a priori} (the method by which they are identified is outside the scope of this study). This simplifies the problem as only said robots execute their filter, which leads to the performance shown in \Cref{fig:BRAVOFalse_dtsd_den1_noP100}.

Apart from when the environment is ambiguous and the true accuracy is low, there is a remarkable improvement in the estimation performance by robots with the \ASDFm{}. This is most noticeable at the lowest nominal filtering period (\Cref{factor:lower_t_better}). At $\tau = 1000$, the \ASDFm{} drastically reduces performance degradation---even preventing any decline in some cases---with the increase of flawed robot proportion. As the nominal filtering period lengthens, the magnitude of degradation mitigated reduces, though estimation performance with the filter is \emph{better} than without it in general. (At $f = 0.95$, the score difference for $50\%$-flawed swarms without filtering and filtering at $\tau=4000$ is not discernible.) This suggests that it may be beneficial to execute the filter more frequently should one value the consistency of estimate performance over computational efficiency.

As for when $f = 0.55$ and $b = 0.55$ (\Cref{fig:BRAVOFalse_dtsd_den1_noP100}, top left), running the \ASDFm{} at $\tau = 1000$ maintains the performance level set by a flawless swarm ($P = 0\%$). However, the effects of the \Cref{factor:specific_f_bt_combo,factor:lower_ba_volatile} undermine the filter's effort to boost estimation accuracy as $P$ increases. Much like the baseline case of the same favorable $f$ and $b$ combination, estimating the fill ratio with a higher $\hat{b}$ enables the swarm to converge on an estimate faster without sacrificing much on accuracy. The \ASDFm{} `interferes' with this process by reducing $\hat{b}$ closer to $b$, thus slowing down convergence. With a lower $\tau$, the convergence decelerates further, leading to the case where $\tau = 1000$ underperforming that at $\tau = 4000$.

\begin{figure*}[!t]
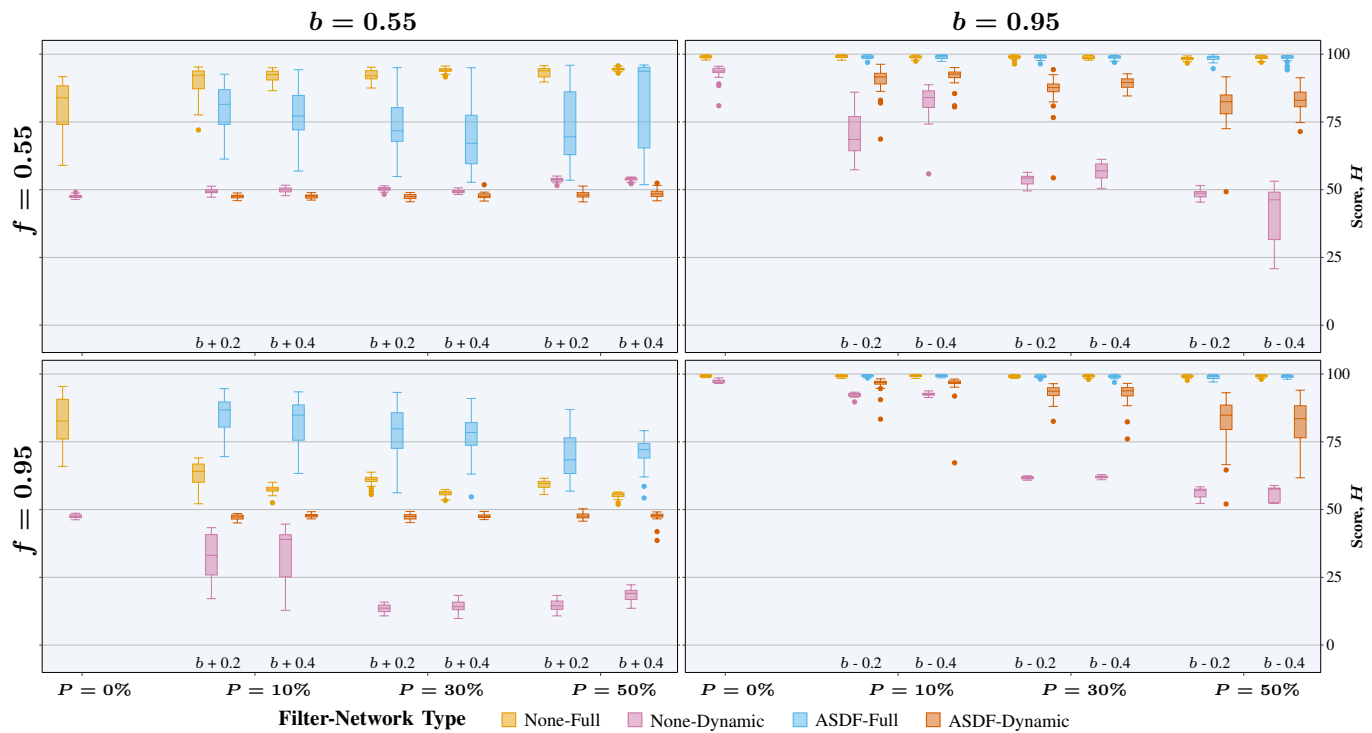
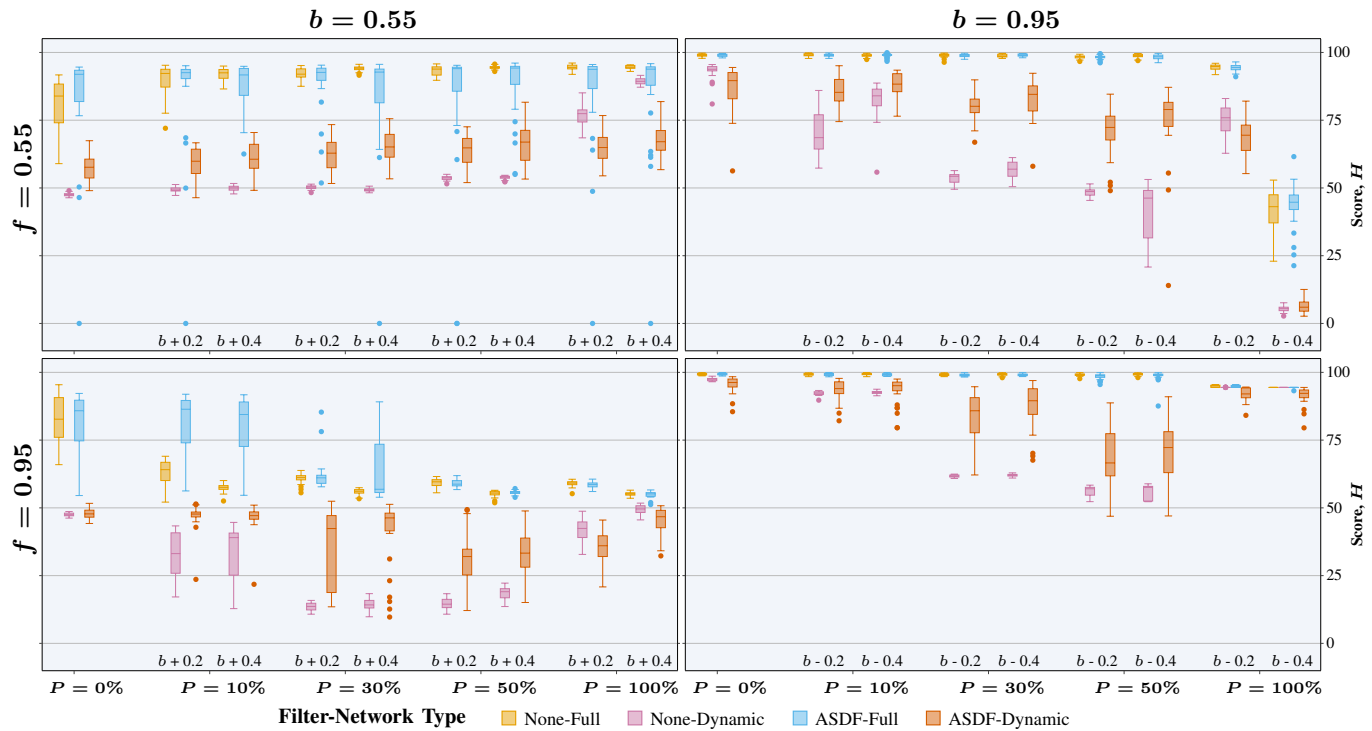

    \centering
    \begin{subfigure}{\textwidth}
        \centering
        \def\svgwidth{\textwidth}
        
        \caption{Scores when \emph{only flawed robots} execute the ASDF. As expected, the fully connected robots are much better in informed estimation, though in some cases the dynamic robots fare quite well (underconfident robots at $P = 10\%$).}
        \label{fig:BRAVOFalse_stsd_v_dtsd_den1_commsp1}
    \end{subfigure}
    \par \bigskip
    \begin{subfigure}{\textwidth}
        \centering
        \def\svgwidth{\textwidth}
        
        \caption{Scores when \emph{all robots} execute the ASDF. Similar to (a), some of the dynamic robot performances come close to those of fully connected robots, albeit with a wider spread.}
        \label{fig:BRAVOTrue_stsd_v_dtsd_den1_commsp1}
    \end{subfigure}
    \caption{Comparison of informed estimation performance between fully connected and dynamic robots (box colors), where each box represents 30 trial scores, $H$, from \Cref{eq:overall_score}. Higher $H$ is preferred as it indicates better performance. Scores of experiments with different flawed assumed accuracies $\hat{b}$ are shown (secondary $x$-axis) as deviations from $b$, plotted for each $P$ (primary $x$-axis). At $P = 0\%$, all robots have correct assumed accuracies, hence the absence of a secondary $x$-axis label. The \textit{No Filtering} data is identical in both (a) and (b). Both sets of scores are computed with $K_{max} = 40000$, $e_{max} = 0.45$.}
    \label{fig:BRAVO_stsd_v_dtsd_den1_commsp1}
\end{figure*}

\paragraph{Fully connected robots}
Much like in the baseline case, a fully connected swarm with the \ASDFm{} achieves a much better estimation performance than the dynamic swarm, as a consequence of the higher information exchange rate (\Cref{fig:BRAVOFalse_stsd_v_dtsd_den1_commsp1}). The trends between the no-filtering and \ASDFm{} cases are similar as well, barring the experiments with underconfident robots (right column). This is where performance is almost perfect even without filtering and its cause was addressed in \Cref{sec:simulated_baseline_performance}.

So far, by identifying flawed robots to execute the filter, the estimation performance by swarms with the \ASDFm{} outperforms the baseline. Techniques to identify flawed robots, however, can be challenging to design and implement. A centralized approach may even be required for such techniques to succeed, compromising the benefits that a decentralized swarm brings. A more reasonable approach is to instead allow all robots, flawed and not, to run the filter and rely on the adaptive activation mechanism for the execution decision.

% Combined score boxplot: STSD vs DTSD P = 0 to 100%

\subsection{\ASDFp{} Performance: Flawed Robot Identities are Unknown}
\label{sec:simulated_asdf_perf_true}

\paragraph{Dynamic robots}
Without prior knowledge of which robots are flawed, we studied a swarm's estimation capabilities without the targeted execution of the ASDF (\Cref{fig:BRAVOTrue_dtsd_den1}).

Intuitively, when all the robots are able to modify their assumed accuracy, flawed robots do not have a reference value to adhere to (indirectly via communicating $\hat{x}$ with correct robots). The possibility that the inaccurate $\hat{b}$ can be corrected diminishes, particularly if the flawed robots are no longer a minority. Conversely, correct robots could also be misguided to modify their (initially) accurate $\hat{b}$, submitting their swarm to conform to an inaccurate estimate. Hence, we see a wider spread of estimation performances by the \ASDFp{} when compared to the \ASDFm{} (\Cref{fig:BRAVO_dtsd_den1}), where correct robots serve as assumed accuracy `anchors'.

Similar to the \ASDFm{}, the \ASDFp{} significantly enhances the estimation performance over the no-filtering swarm, albeit to a lesser degree. Alongside a higher variance between trials, the \ASDFp{} is moderately worse than the \ASDFm{}; performance is undeniably inferior when the proportion of flawed robots is high ($P = 50\%$). This is understandable because of the uncertainty in the correct robots' assumed accuracy. The only exception is when $f = 0.55$, $b = 0.55$ (\Cref{fig:BRAVOTrue_dtsd_den1}, top left), much like in the baseline case where the \Cref{factor:specific_f_bt_combo,factor:lower_ba_volatile,factor:higher_ba_overrides} are at play. Here the median score is consistently higher than the maximum score of the \ASDFm{} experiments. This result is due to the ubiquitous execution of the \ASDFp{}, as correct robots may mistakenly adjust their correct assumed accuracy to an incorrect one. Naturally, this is detrimental to the estimate accuracy, but the decline is minor as opposed to the gain in convergence speed, as shown in \Cref{fig:BRAVO_sep_dtsd_den1}. The scores for the \ASDFp{} estimate spread out horizontally with minimal vertical movement as compared to the \ASDFm{} scores. Beyond that, the estimation performance as a whole rises steadily with the increase of $P$ due to the influence of the \Cref{factor:specific_f_bt_combo}.

We also note an interesting phenomenon in \Cref{fig:BRAVO_sep_dtsd_den1} at $P = 50\%$ with the \ASDFp{}. The scores for the overconfident robots at $f = 0.95$ spread out in a speed-accuracy tradeoff curve, but not for the underconfident robots, nor for both cases at $f = 0.55$. To understand this, first consider that only the \emph{overconfident robots at $f = 0.95$} and the \emph{underconfident robots at $f = 0.55$} are anticipated to have similar outcomes, due to the symmetric local estimate errors (\Cref{tab:prob_obs_black_and_local_est}). (The remaining two cases, underconfident robots at $f=0.95$ and overconfident robots at $f=0.55$, have highly accurate local estimates regardless of $\hat{b}$ due to the \Cref{factor:specific_f_bt_combo}.) Then, due to the \Cref{factor:higher_ba_overrides}, the odds of overconfident robots misleading the correct robots increases---particularly since the flawed robots are no longer a minority, leading to the speed-accuracy tradeoff spread seen in the case with the unambiguous environment. As for underconfident robots in an ambiguous environment, the \Cref{factor:higher_ba_overrides} works \emph{against} the flawed robots by making their local estimates less influential.

That being said, using the filter has no benefit to an entire swarm of flawed robots (\ASDFm{} is the same as \ASDFp{} here). As shown in \Cref{fig:BRAVOTrue_dtsd_den1}, the filter's performance fails to surpass baseline results in all $P = 100\%$ cases. At best, the ASDF maintains the estimation performance but is mostly worse than the baseline. The inferiority comes solely from lower ASDF convergence scores as compared to the no-filtering case, observed by comparing their separate score plots (\Cref{fig:no_filtering_sep_dtsd_den1,fig:BRAVO_sep_dtsd_den1}). This suggests that the homogeneity in the robots' assumed accuracy makes it more likely that a robot's local estimate would be similar to its neighbors', effectively accelerating convergence on the informed estimate.

\begin{figure}[!t]
    \centering
    \def\svgwidth{\columnwidth}
    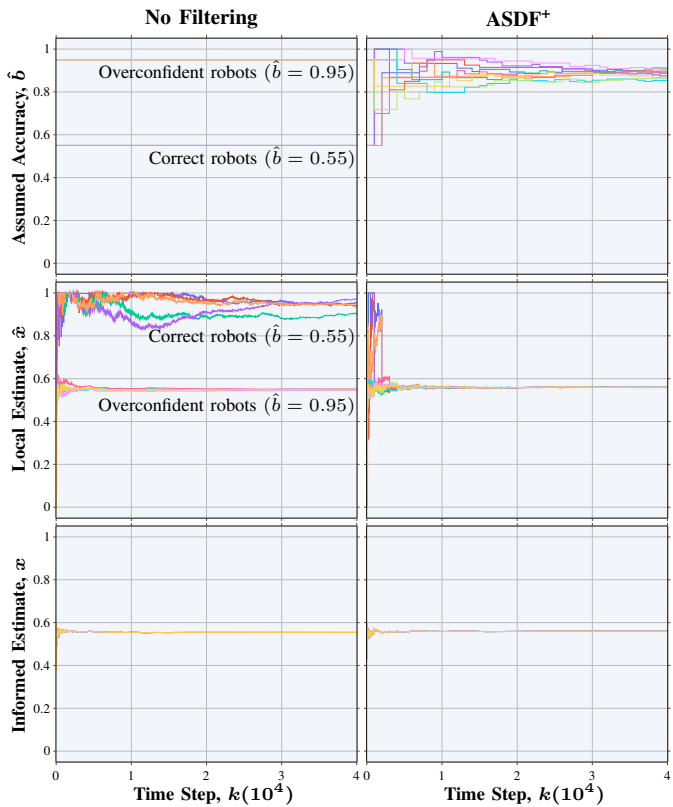
    \caption{Time series trajectories of $\hat{b}$, $\hat{x}$, and $x$ for the no-filtering (left column) and the \ASDFp{} (right column) swarms at $f=0.95$, $P = 50\%$, and $\tau = 1000$, obtained from a single trial. In both cases, the robots are fully connected and have a true accuracy of $b = 0.55$. The \Cref{factor:higher_ba_overrides} affects both swarms, albeit in different ways. For the no-filtering robots, the correct robots' local estimates are suppressed by those of the overconfident robots. For the \ASDFp{} robots, the correct robots' assumed accuracy is corrupted by the overconfident robots.}
    \label{fig:stsd_corfilt1_timeseries_comparison}
\end{figure}

\paragraph{Fully connected robots}
The estimation performance by the fully connected robots is overwhelmingly better than that of the dynamic robots (\Cref{fig:BRAVOTrue_stsd_v_dtsd_den1_commsp1}), consistent with the baseline and \ASDFm{} results. The trend between the no-filtering and \ASDFp{} performance, however, is quite different when compared to the case of the dynamic robots. We covered reasons for the deviation of the underconfident robots in \Cref{sec:simulated_baseline_performance}, but here even the overconfident robots behave differently (albeit when $f = 0.95$ the trend is consistent for $P \leq 10\%$). In general, fully connected swarms with overconfident robots that run the \ASDFp{} \emph{do not} improve upon their baseline performance.

The discrepancy in the ambiguous environment setting ($f = 0.55$) is the result of a combination of factors: the \Cref{factor:specific_f_bt_combo} ensures that informed estimation is quite accurate to begin with (regardless of $\hat{b}$), while the \Cref{factor:lower_ba_volatile,factor:higher_ba_overrides} delay convergence with the smaller number of overconfident robots. In fact, the \Cref{factor:higher_ba_overrides} plays a vital role in \emph{grounding} the estimates because $\alpha$ values from overconfident robots are orders of magnitude higher than the correct robots (\Cref{fig:no_filter_stsd_alpha_comparison}). Fewer overconfident robots thus result in a situation where most robots have comparable $\alpha$ values which \emph{weakens the grounding effect}, ultimately prolonging the estimates' settling time.

On the other hand, the ineffective filtering performance in the unambiguous environment setting ($f = 0.95$) is dominated by the \Cref{factor:higher_ba_overrides}. With a harder $f$ and $b$ combination (\Cref{tab:prob_obs_black_and_local_est}), the overconfident robots' local estimates start off being highly inaccurate, corrupting correct robots' (social and informed) estimates easily due to the higher $\alpha$ values. This creates a positive feedback loop for $\hat{b}$ as it is modified by the incorrect social estimate, which in turn reinforces the incorrect local estimate. Eventually, the informed estimates of the entire swarm become inaccurate, as shown in \Cref{fig:stsd_corfilt1_timeseries_comparison}. That said, this can still be mitigated as long as only a tiny fraction of the swarm is flawed ($P \leq 10\%$).

% \begin{figure}[!t]
%     \centering
%     \includegraphics[width=0.95\columnwidth]{figures/no_filter_vs_BRAVOTrue_stsd_p0p10_tfr950_flwb950_corb550.pdf}
%     \caption{nofilter v bravo true stsd p0 p10 tfr950 flwb950 corb550}
%     \label{fig:blah}
% \end{figure}

In comparing the \ASDFp{} against the \ASDFm{}, we see the wider performance spread analogous to the dynamic topology experiments. Their performance levels are also similar, apart from the overconfident robots at higher $P$ when $f = 0.95$. As previously discussed, the \Cref{factor:higher_ba_overrides} tempers the benefit of the filter in the \ASDFp{} case, but here it is less effectual when correct assumed accuracies are unmovable in the \ASDFm{} case. For this reason, we continue to see the \ASDFm{} improving upon baseline performance even with more flawed robots.

Overall, regardless of the communication topology, the ASDF achieves its purpose in improving informed estimation when assumed accuracies may be flawed. Under typical operating conditions, where sensor degradation causes some members of the swarm to become overconfident ($\hat{b} > b$), the \ASDFm{} is able to maintain the baseline performance of a flawless swarm. Even without the targeted approach to filter execution, estimation performance with the \ASDFp{} is better than without it. Alternately, when some robots are miscalibrated to a lower assumed accuracy ($\hat{b} < b$), the ASDF can significantly improve the estimation performance of the same swarm without the filter. If the \emph{entire swarm} has a flawed assumed accuracy, however, the advantage of using the ASDF vanishes; a solution for such a situation may prove to be infeasible for a minimalistic swarm without additional assumptions.
\section{Physical Experiments}
\label{sec:physical_experiments}

\begin{figure*}[t]
    \centering
    \begin{subfigure}{0.49\textwidth}
        \centering
        \includegraphics[width=\textwidth]{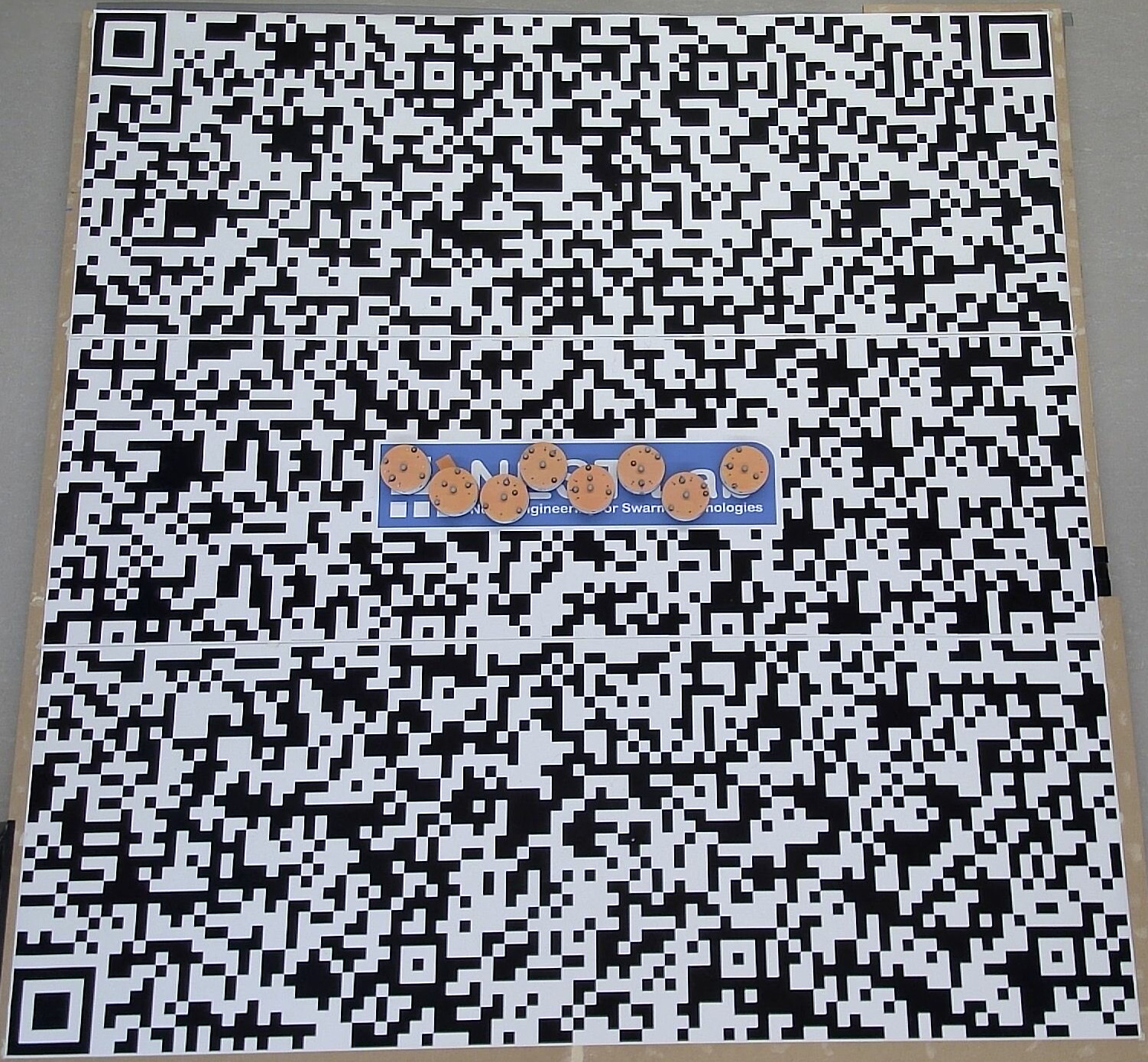}
    \end{subfigure}
    \begin{subfigure}{0.49\textwidth}
        \centering
        \includegraphics[width=\textwidth]{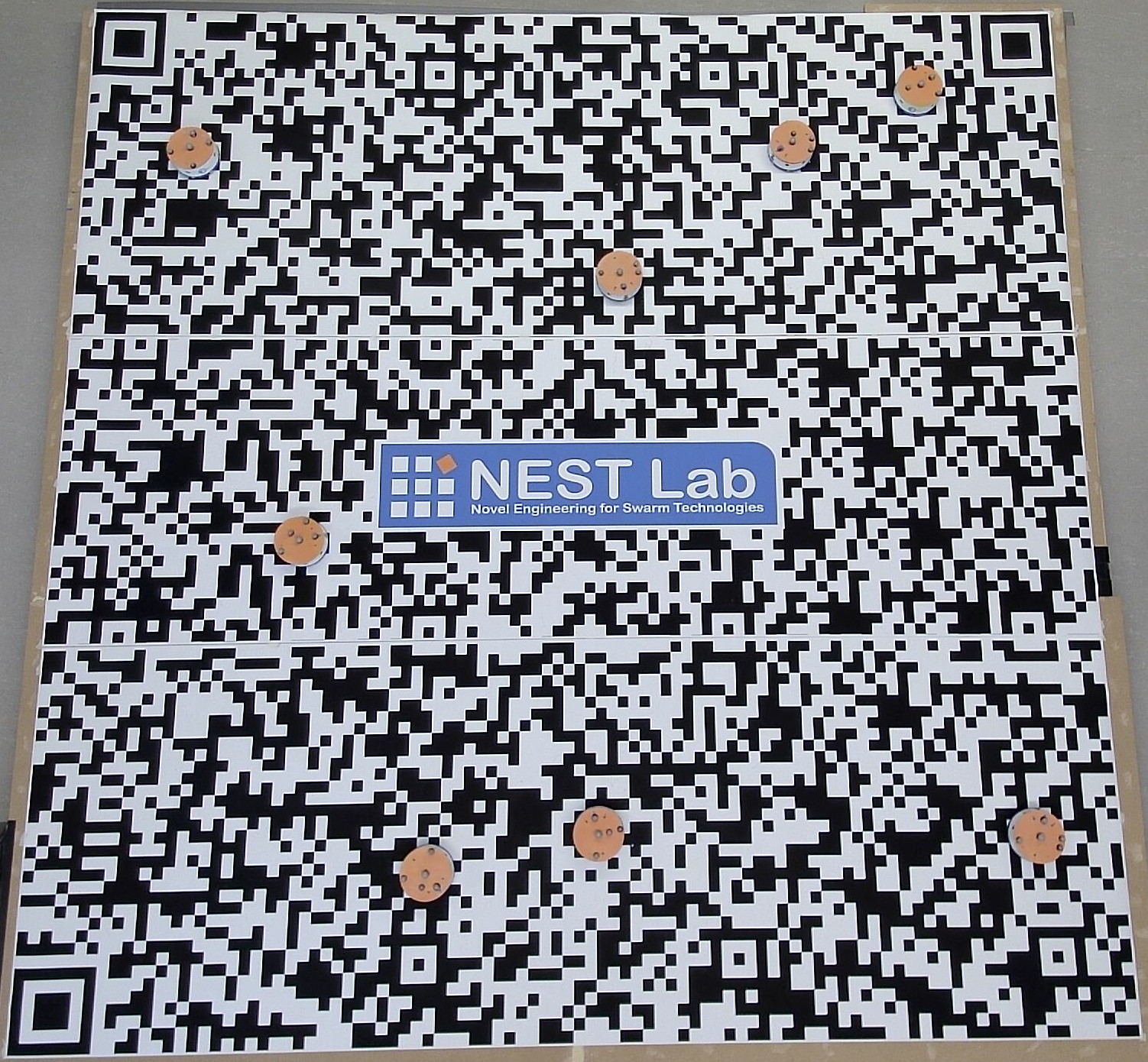}
    \end{subfigure}
    \caption{Our physical experiment setup of a single random trial. The left image shows the robots' starting positions while the right image shows the diffused robots after the trial has run for a while. The logo in the middle of the QR code is perceived as non-black by the robots' ground sensor.}
    \label{fig:physical_setup}
\end{figure*}

\subsection{General Setup}
We run our physical experiments with $N = 8$ Khepera IV robots on a $\unit[3.0353 \times 3.0226]{m^2}$ platform. The robots diffuse across a QR code surface, as shown in \Cref{fig:physical_setup}. With a communication radius of $\unit[0.7]{m}$, the swarm density is $D \approx 1.3$. The QR code has a fill ratio of $f = 0.448$, which makes the physical environment an ambiguous one. Each robot possesses 4 perfect ground sensors that can detect whether a surface is black or not; for simplicity, our robots utilize only their front right sensor to make observations.

Differently from the simulated experiments, we perform each trial of the physical experiments for $1000$ seconds (equivalent to $10,000$ simulated time steps) due to logistical reasons. We run 3 trials for each combination of the following experimental parameters for both the baseline (no-filtering) and \ASDFp{} ($\tau = 1000$) cases:
\begin{itemize}
    \item flawed robot proportions, $P = \{ 0, 25, 50, 100 \} \%$;
    \item flawed assumed accuracies, $\hat{b} = \{ 0.55, 0.75, 0.95 \}$.
\end{itemize}
This amounts to a collection of over 16 hours of experimental data. The true accuracies for correct robots are assigned to $b = 0.99$ to prevent numerical errors; essentially, the flawed robots in these experiments are underconfident. Each robot travels at a forward speed of $\unit[0.1]{ms^{-1}}$ with a $\unit[2]{Hz}$ observation rate. This accounts for the size of an atomic QR code tile which is $\unit[4.5]{cm}$ long diagonally. For the \ASDFp{} experiments, we use a type II error probability of $\omega = 0.05$.

\subsection{Baseline vs. \ASDFp{} Performance}
\begin{figure*}[!t]
    \centering
    \begin{subfigure}{\textwidth}
        \centering
        \def\svgwidth{0.8\textwidth}
        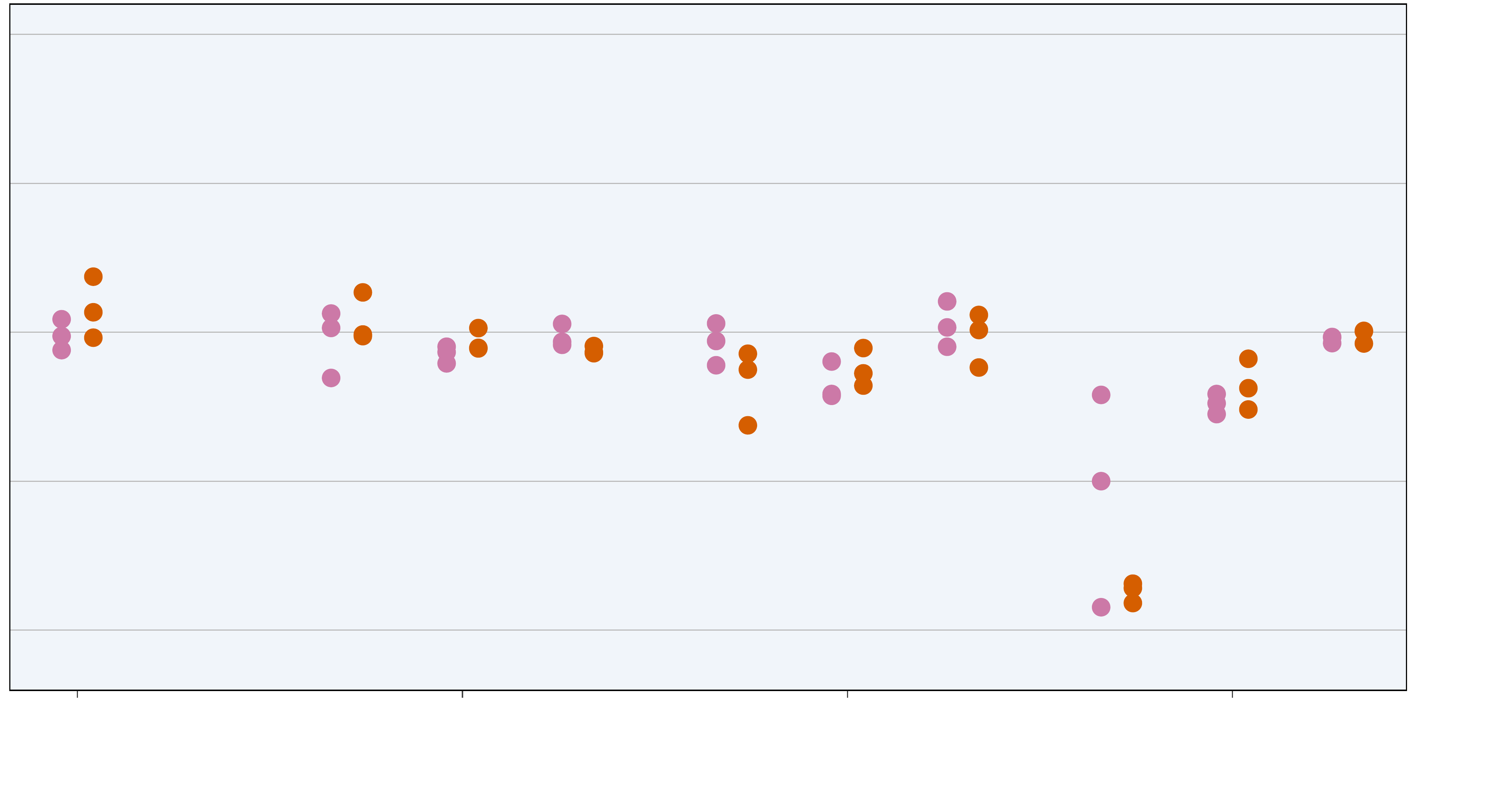
        \caption{Comparison of informed estimation performance between the no-filtering and \ASDFp{} swarms, where both have comparable scores $H$ (higher $H$ is preferable). This is similar to the box plots in \Cref{fig:BRAVO_dtsd_den1}, except the boxes were removed here due to the fewer samples. Each point represents a single experiment trial, plotted across different assumed accuracies $\hat{b}$ (secondary $x$-axis) and flawed robot proportions $P$ (primary $x$-axis). At $P = 0\%$, all robots correct assumed accuracies, hence the absence of a secondary $x$-axis label.}
        \label{fig:BRAVOTrue_dtsd_physical_tfr0_den1_corb550}
    \end{subfigure}
    \par \bigskip
    \begin{subfigure}{\textwidth}
        \centering
        \def\svgwidth{0.85\textwidth}
        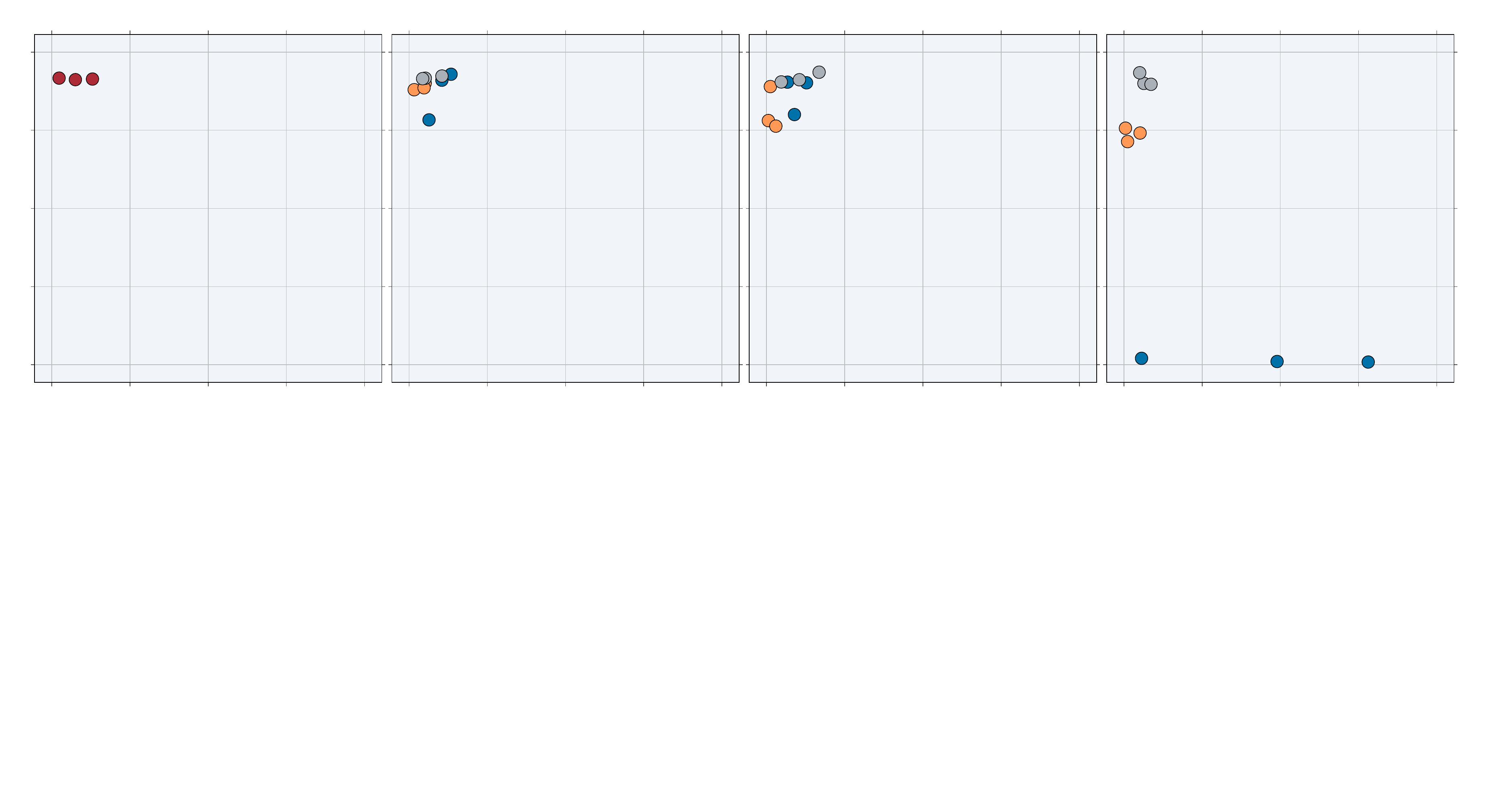
        \caption{Comparison of informed estimation performance in terms of convergence and accuracy between the no-filtering and \ASDFp{} swarms. In both cases, their convergence ($h_K$) scores and accuracy ($h_e$) scores are comparable, except at $P = 100\%$; higher $h_K$ (rightward) and $h_e$ (upward) scores are preferred. This is similar to the scatter plots in \Cref{fig:sep_dtsd_den1}. Each point represents the $h_K$ and $h_e$ scores (from \Cref{eq:consensus_scaling}) for a single trial.}
        \label{fig:no_filter_v_BRAVOTrue_sep_dtsd_physical}
    \end{subfigure}
    \caption{Performance of the no-filtering and \ASDFp{} robots from the physical experiments. The scores are computed with $K_{max} = 10000$, $e_{max} = 0.45$.}
    \label{fig:no_filter_v_BRAVOTrue_dtsd_physical}
\end{figure*}

With a fill ratio of $f = 0.448$ and a true accuracy of $b = 0.99$, we expected the physical experiments to be comparable with the simulated case of $f = 0.55$ and $b = 0.95$, as shown in the top left plot of \Cref{fig:BRAVOTrue_dtsd_den1}. However, we see that the filter usage has no effect on the estimation performance with respect to a swarm with no filtering (\Cref{fig:BRAVOTrue_dtsd_physical_tfr0_den1_corb550}). This is true across all proportions of flawed robots $P$ and flawed assumed accuracies $\hat{b}$ that we tested, bar a slight difference in the $P = 100\%$, $\hat{b} = 0.55$ case. By comparing the convergence and accuracy scores separately in \Cref{fig:no_filter_v_BRAVOTrue_sep_dtsd_physical}, we see that this minor distinction is due to the improved convergence performance, but accuracy levels are similar. In fact, accuracy scores are quite high for all the baseline and \ASDFp{} cases, but convergence scores are poor.

We attribute the cause of the ineffectual filter usage to the compressed duration in the physical experiments. Given a shorter time frame, it makes sense that the robots' estimates remain quite volatile by the end of the trial. Additionally, the observation rate of $\unit[2]{Hz}$ means that the robots collect readings at one-fifth the speed done in simulated experiments. Effectively, the amount of observations $t$ made by the robots in the physical experiments is $20$ times fewer than those in the simulated experiments. This is why the robots from the physical experiments have a dismal convergence score relative to their simulated counterparts. We confirm this visually using \Cref{fig:BRAVO_sep_dtsd_den1}, where the underconfident robots' convergence scores at $f = 0.55$ are significantly higher. It is also reasonable to ascribe the largely identical performance across increasing $P$ to a reduced $t$, as the physical experiments were not run long enough for the estimates to reach an expected convergence point.

Nonetheless, these results show that the application of the \ASDFp{} does not harm a swarm's estimation performance in the short term. Then, as the swarm operates for longer durations, it becomes increasingly likely that the estimate convergence improves \emph{with} filter usage than without, as exhibited in the simulated results. Eventually, \Cref{fig:BRAVOTrue_dtsd_den1} show that the overall performance gain offered by the ASDF is evident.

\section{Conclusions}
\label{sec:conclusions}

In this work, we analyzed the swarm's ability to perform collective perception when the robots' true sensor accuracies are wrongly assigned for the MCP algorithm. The overall estimation performance suffers substantially when robots have an incorrect assumed accuracy, though by pure chance an ambiguous environment allows overconfident robots to thrive over a flawless swarm. To subdue the drop in performance, we proposed the ASDF: it is a minimal algorithm---suitable for computationally constrained swarm robots---and leverages information from a robot's neighbors to update said robot's assumed accuracy.

Our efforts yield noteworthy results: estimation performance has a milder decline, even retaining the levels produced by flawless swarms in certain conditions. Notably, when only flawed robots execute the filter, swarms that constitute at most $50\%$ overconfident robots achieve performance on par with that of flawless robots. Moreover, the benefits of the ASDF can be reaped regardless if flawed robot identities are known; even without a targeted execution of the filter, estimation performance is still superior to the no-filtering baseline. That said, the filter is inadequate when the entire swarm is flawed; in such situations, a more drastic measure may be required. We also found that a shorter-term usage of the filter has no harmful effects on the estimation performance; this is a useful attribute as it implies that a minimum operating duration does not limit ASDF-enabled swarms.

As part of our future work, we intend to study the effectiveness of the MCP algorithm under conditions where the true sensor accuracy may degrade over time. We also aspire to update the assumed sensor accuracy so that it correctly reflects its time-varying, true counterpart.
\section{Acknowledgements}
\noindent This research was performed using computational resources supported by the Academic \& Research Computing group at Worcester Polytechnic Institute. This work was partially supported by grant \#W911NF2220001.

\appendices
\section{Asymptotic Normality of MLEs} \label[appdx]{appdx:asymp_norm}
Let $X_1, \ldots, X_n$ be \textit{i.i.d.} random variables with a probability density function parametrized by $\theta$: $p(x \mid \theta)$. The MLE $\hat{\theta}$ is found as the solution to
\begin{align*}
    \frac{\partial}{\partial \theta} \ln p(X_1, \ldots, X_n \mid \theta) = 0
\end{align*}
where $p(X_1, \ldots, X_n \mid \theta)$ is the likelihood function. The proof that the limiting distribution of an MLE is Gaussian depends on the following theorems \cite{casellaStatisticalInference2002}.

\begin{theorem}[Asymptotic efficiency of MLEs]\label{thm:asymp_eff_mle}
    Let $\hat{\theta}$ denote the MLE of $\theta$. We assume the following regularity conditions on $p(x \mid \theta)$:
    \begin{enumerate}
        \item $X_1, \ldots, X_n \overset{iid}{\sim} p(x \mid \theta)$, $x \in \mathcal{X}$.
        \item $\theta$ is identifiable; that is, if $\theta \neq \theta^\prime$, then $p(x \mid \theta) \neq p(x \mid \theta^\prime)$.
        \item The densities $p(x \mid \theta)$ have common support and $p(x \mid \theta)$ is differentiable in $\theta$.
        \item The parameter space $\Omega$ contains an open set $\omega$ of which the true parameter value $\theta_0$ is an interior point.
        \item $\forall x \in \mathcal{X}$, $p(x \mid \theta)$ is three times differentiable with respect to $\theta$, the third derivative is continuous in $\theta$, and $\int p(x \mid \theta) dx$ can be differentiated three times under the integral sign.
        \item For any $\theta_0 \in \Omega$, there exists a positive number $c$ and a function $M(x)$ such that
              \begin{gather*}
                  \bigg| \frac{\partial^3}{\partial \theta^3} \ln p(x \mid \theta) \bigg| \leq M(x) \\
                  \forall x \in \mathcal{X}, \quad \theta_0 - c < \theta < \theta_0 + c,
              \end{gather*}
              with $\mathbb{E}_{\theta_0}[M(X)] < \infty$.
    \end{enumerate}
    Then, $\hat{\theta}$ is a consistent and asymptotically efficient estimator of $\theta$. That is, $\sqrt{n}(\hat{\theta} - \theta)$ converges in distribution to a zero-mean Gaussian distributed variable with an asymptotic variance equal to the Cram\'er-Rao lower bound:
    \begin{equation*}
        \sqrt{n} (\hat{\theta} - \theta) \overset{d}{\rightarrow} G, \quad G \sim \mathcal{N}(0, \nicefrac{1}{\mathcal{I}(\theta)}).
    \end{equation*}
\end{theorem}

\begin{theorem}[Continuous mapping theorem]\label{thm:cont_map}
    Let the sequence $\{ Y_1, Y_2, \ldots, Y_n \}$ converge to a random variable $Y$ in distribution and let $h$ be a continuous function. Then $\{ h(Y_1), h(Y_2), \ldots, h(Y_n) \}$ converges to $h(Y)$ in distribution.
\end{theorem}

We show the asymptotic normality of the MLE $\hat{\theta}$ by applying \Cref{thm:cont_map} to the result of \Cref{thm:asymp_eff_mle}.
\begin{proof}
    \begin{align*}
        \hat{\theta} & = \frac{1}{\sqrt{n}} \big( \sqrt{n}(\hat{\theta} - \theta) \big) + \theta \\
                     & = h(\sqrt{n}(\hat{\theta} - \theta))                                      \\
                     & \overset{d}{\rightarrow} h(G)                                             \\
                     & = \frac{1}{\sqrt{n}} G + \theta                                           \\
                     & \coloneq G^\prime, \text{ where }
        G^\prime \sim \mathcal{N}(\theta, \nicefrac{1}{n\mathcal{I}(\theta)}).
    \end{align*}
    % where
    % $G^\prime$ is a Gaussian distributed random variable with a mean of $\theta$ and a variance of $\nicefrac{1}{n\mathcal{I}(\theta)}$.
\end{proof}
% \section{Details of the Likelihood Ratio Test for Adaptive Filter Activation} \label[appdx]{appdx:lrt_derivation}
\section{Derivation of the Likelihood Ratio Test for Adaptive Filter Activation} \label[appdx]{appdx:lrt_derivation}

% \paragraph{Derivation of the likelihood ratio test}
To decide whether an update to its assumed sensor accuracy $\hat{b}_i$ is needed, robot $i$ compares its local estimate $\hat{x}_i$ with those received from its neighbors, $\hat{x}_j$. If $\hat{x}_i$ is close enough to $\hat{x}_j$---implying that $\hat{b}_i$ is good enough to generate estimates consistent with others---then no modification is required. Conversely, if $\hat{x}_i$ is substantially different from $\hat{x}_j$---indicating an assumed sensor accuracy irregularity among the robots involved---then an update is necessary. Note that this assumes that the majority of the neighbors have a correct assumed accuracy.

In formalizing this decision-making process into a hypothesis test, we assume that $\hat{x}_j$ follows a Gaussian distribution with an unknown mean $\mu$ and an unknown variance $\sigma^2$. (The validity of this assumption is shown in \Cref{appdx:asymp_norm}.) As robot $i$, our hypotheses are stated as follows.
\begin{equation*}
    H_0: \mu = \hat{x}_i, \qquad H_1: \mu \neq \hat{x}_i
\end{equation*}
where the rejection of the null hypothesis leads to activation of the filter.

Our test is generated using the ratio of the null ($H_0$) likelihood to the maximum likelihood, \textit{i.e.,} the likelihood ratio test (LRT). The LRT statistic is defined as
\begin{align*}
    \lambda(\hat{x}_j) & \coloneq \frac{\sup_{\mu = \hat{x}_i, \sigma^2 \in \mathbb{R}^+} L(\mu, \sigma^2 \mid \hat{x}_j)}{\sup_{\mu \in \mathbb{R}, \sigma^2 \in \mathbb{R}^+} L(\mu, \sigma^2 \mid \hat{x}_j)}
\end{align*}
where $L(\mu, \sigma^2 \mid \hat{x}_j)$ is the likelihood function based on the `evidence', $\hat{x}_j$. In the denominator, the mean $\mu$ and variance $\sigma^2$ that parametrizes the maximum likelihood are the respective estimators:
\begin{align*}
    \mu_{MLE} = \bar{X}_i = \frac{\sum_j^{m_i} \hat{x}_j}{m_i},\quad \sigma^2_{MLE} = \frac{\sum_j^{m_i} (\hat{x}_j - \bar{X}_i)^2}{m_i}
\end{align*}
where $m_i \geq 2$ is the amount of robot $i$'s neighbors. (Note that while $\mu_{MLE}$ is the sample mean, $\sigma^2_{MLE}$ is \emph{not} the sample variance.) Meanwhile, under the null region, the restricted supremum leads to a likelihood parametrized by $\mu = \hat{x}_i$ and $\sigma^2 = \sigma^2_0$. The latter is the mean squared deviation from $\mu = \hat{x}_i$:
\begin{align*}
    \sigma^2_0 = \frac{\sum_j^{m_i} (\hat{x}_j - \hat{x}_i)^2}{m_i}.
\end{align*}

By substituting the parameter estimators and recalling that the likelihoods are Gaussian-distributed joint probability density functions, we reduce the LRT statistic:
\begin{align*}
    \lambda(\hat{x}_j) & = \frac{L(\hat{x}_i, \sigma^2_0 \mid \hat{x}_j)}{ L(\mu_{MLE}, \sigma^2_{MLE} \mid \hat{x}_j)}                                                                                                                                                                                                                                                                                                  \\
                       & = \dfrac{ \bigg( \dfrac{1}{\sqrt{2 \pi \sigma^2_0}} \bigg)^{m_i} \exp \bigg\{ -\dfrac{\sum_j^{m_i} (\hat{x}_j - \hat{x}_i)^2}{2\sigma^2_0} \bigg\} }{\bigg( \dfrac{1}{\sqrt{2 \pi \sigma^2_{MLE}}} \bigg)^{m_i} \exp \bigg\{ -\dfrac{\sum_j^{m_i} (\hat{x}_j - \mu_{MLE})^2}{2\sigma^2_{MLE}} \bigg\} }                                                                                         \\[5pt]
                       & = \dfrac{ \bigg( \dfrac{1}{\sqrt{2 \pi \sigma^2_0}} \bigg)^{m_i} \exp \bigg\{ -\dfrac{\sum_j^{m_i} (\hat{x}_j - \hat{x}_i)^2}{2(\nicefrac{\sum_j^{m_i}(\hat{x}_j - \hat{x}_i)^2}{m_i}) } \bigg\} }{\bigg( \dfrac{1}{\sqrt{2 \pi \sigma^2_{MLE}}} \bigg)^{m_i} \exp \bigg\{ -\dfrac{\sum_j^{m_i} (\hat{x}_j - \bar{X}_i)^2}{2(\nicefrac{\sum_j^{m_i}(\hat{x}_j - \bar{X}_i)^2}{m_i}) } \bigg\} } \\[5pt]
                       & = \bigg( \sqrt{\dfrac{\sigma^2_{MLE}}{\sigma^2_0}} \bigg)^{m_i}                                                                                                                                                                                                                                                                                                                                 \\[5pt]
                       & = \bigg( \dfrac{\sum_j^{m_i} (\hat{x}_j - \bar{X}_i)^2}{\sum_j^{m_i} (\hat{x}_j - \bar{X}_i)^2 + m_i(\bar{X}_i - \hat{x}_i)} \bigg)^{\nicefrac{m_i}{2}}.
\end{align*}
Subsequently, we specify our null hypothesis rejection region by assessing if the LRT statistic is less than the threshold $0 < a < 1$:
\begin{align*}
    R_i & = \{\hat{x}_j: \lambda(\hat{x}_j) < a \}                                                                                                                                                \\[5pt]
        & = \bigg\{ \hat{x}_j: \bigg( \dfrac{\sum_j^{m_i} (\hat{x}_j - \bar{X}_i)^2}{\sum_j^{m_i} (\hat{x}_j - \bar{X}_i)^2 + m_i(\bar{X}_i - \hat{x}_i)} \bigg)^{\nicefrac{m_i}{2}} < a \bigg\}.
\end{align*}
For computational convenience, we manipulate the inequality to obtain a more familiar test statistic:
\begin{align*}
    R_i & = \bigg\{ \hat{x}_j: \bigg( \dfrac{\sum_j^{m_i} (\hat{x}_j - \bar{X}_i)^2}{\sum_j^{m_i} (\hat{x}_j - \bar{X}_i)^2 + m_i(\bar{X}_i - \hat{x}_i)} \bigg)^{\nicefrac{m_i}{2}} < a \bigg\} \\[5pt]
        & = \bigg\{ \hat{x}_j: \dfrac{m_i(\bar{X}_i - \hat{x}_i)}{\sum_j^{m_i} (\hat{x}_j - \bar{X}_i)^2} > a^{\nicefrac{-2}{m_i}} - 1 \bigg\}                                                   \\[5pt]
        & = \bigg\{ \hat{x}_j: \dfrac{\bar{X}_i - \hat{x}_i}{\nicefrac{S_i^2}{m_i}} > (a^{\nicefrac{-2}{m_i}} - 1)(m_i - 1) > 0 \bigg\}                                                          \\[5pt]
        & = \bigg\{ \hat{x}_j: \dfrac{|\bar{X}_i - \hat{x}_i|}{\nicefrac{S_i}{\sqrt{m_i}}} > \sqrt{(a^{\nicefrac{-2}{m_i}} - 1)(m_i - 1)} \eqcolon c \bigg\}                                     \\[5pt]
\end{align*}
where $S_i^2 = \sum_j^{m_i} (\hat{x}_j - \bar{X}_i)^2 / (m_i - 1)$ is the sample variance and $c \geq 0$ is the new user-specified threshold.

After manipulation, the resulting test statistic is closely related to the following statistic that follows a Student's $t$ distribution with $m_i - 1$ degrees of freedom:
\begin{align*}
    \frac{\bar{X}_i - \mu}{\nicefrac{S_i}{\sqrt{m_i}}} \sim t_{m_i - 1}.
\end{align*}
We make use of this statistic when determining $c$ through the statistical power function in \Cref{sec:sensor_degradation_filter_activation}.

\balance
\bibliographystyle{IEEEtran}
\bibliography{IEEEabrv,references}

\begin{thebibliography}{10}
\providecommand{\url}[1]{#1}
\csname url@samestyle\endcsname
\providecommand{\newblock}{\relax}
\providecommand{\bibinfo}[2]{#2}
\providecommand{\BIBentrySTDinterwordspacing}{\spaceskip=0pt\relax}
\providecommand{\BIBentryALTinterwordstretchfactor}{4}
\providecommand{\BIBentryALTinterwordspacing}{\spaceskip=\fontdimen2\font plus
\BIBentryALTinterwordstretchfactor\fontdimen3\font minus \fontdimen4\font\relax}
\providecommand{\BIBforeignlanguage}[2]{{%
\expandafter\ifx\csname l@#1\endcsname\relax
\typeout{** WARNING: IEEEtran.bst: No hyphenation pattern has been}%
\typeout{** loaded for the language `#1'. Using the pattern for}%
\typeout{** the default language instead.}%
\else
\language=\csname l@#1\endcsname
\fi
#2}}
\providecommand{\BIBdecl}{\relax}
\BIBdecl

\bibitem{chinMinimalisticCollectivePerception2023}
K.~Y. Chin, Y.~Khaluf, and C.~Pinciroli, ``Minimalistic collective perception with imperfect sensors,'' in \emph{2023 {{IEEE}}/{{RSJ International Conference}} on {{Intelligent Robots}} and {{Systems}} ({{IROS}})}.\hskip 1em plus 0.5em minus 0.4em\relax Detroit, MI, USA: IEEE, Oct. 2023, pp. 8862--8868.

\bibitem{siemensmaCollectiveBayesianDecisionMaking2024}
T.~Siemensma, D.~Chiu, S.~Ramshanker, R.~Nagpal, and B.~Haghighat, ``Collective bayesian decision-making in a swarm of miniaturized robots for surface inspection,'' Apr. 2024.

\bibitem{haghighatApproachBasedParticle2022}
B.~Haghighat, J.~Boghaert, Z.~{Minsky-Primus}, J.~Ebert, F.~Liu, M.~Nisser, A.~Ekblaw, and R.~Nagpal, ``An approach based on particle swarm optimization for inspection of spacecraft hulls by a swarm of miniaturized robots,'' in \emph{Swarm {{Intelligence}}}, M.~Dorigo, H.~Hamann, M.~{L{\'o}pez-Ib{\'a}{\~n}ez}, J.~{Garc{\'i}a-Nieto}, A.~Engelbrecht, C.~Pinciroli, V.~Strobel, and C.~{Camacho-Villal{\'o}n}, Eds.\hskip 1em plus 0.5em minus 0.4em\relax Cham: Springer International Publishing, 2022, vol. 13491, pp. 14--27.

\bibitem{tamjidiUnifyingConsensusCovariance2021}
A.~Tamjidi, R.~Oftadeh, M.~N.~G. Mohamed, D.~Yu, S.~Chakravorty, and D.~Shell, ``Unifying consensus and covariance intersection for efficient distributed state estimation over unreliable networks,'' \emph{IEEE Transactions on Robotics}, vol.~37, no.~5, pp. 1525--1538, Oct. 2021.

\bibitem{majcherczykDistributedDataStorage2021}
N.~Majcherczyk, D.~J. Nallathambi, T.~Antonelli, and C.~Pinciroli, ``Distributed data storage and fusion for collective perception in resource-limited mobile robot swarms,'' \emph{IEEE Robotics and Automation Letters}, vol.~6, no.~3, pp. 5549--5556, Jul. 2021.

\bibitem{valentiniCollectivePerceptionEnvironmental2016}
G.~Valentini, D.~Brambilla, H.~Hamann, and M.~Dorigo, ``Collective perception of environmental features in a robot swarm,'' in \emph{Swarm {{Intelligence}}}, M.~Dorigo, M.~Birattari, X.~Li, M.~{L{\'o}pez-Ib{\'a}{\~n}ez}, K.~Ohkura, C.~Pinciroli, and T.~St{\"u}tzle, Eds.\hskip 1em plus 0.5em minus 0.4em\relax Cham: Springer International Publishing, 2016, vol. 9882, pp. 65--76.

\bibitem{shanDiscreteCollectiveEstimation2021}
Q.~Shan and S.~Mostaghim, ``Discrete collective estimation in swarm robotics with distributed bayesian belief sharing,'' \emph{Swarm Intelligence}, vol.~15, no.~4, pp. 377--402, Dec. 2021.

\bibitem{ebertBayesBotsCollective2020}
J.~T. Ebert, M.~Gauci, F.~{Mallmann-Trenn}, and R.~Nagpal, ``Bayes bots: {{Collective}} bayesian decision-making in decentralized robot swarms,'' in \emph{2020 {{IEEE International Conference}} on {{Robotics}} and {{Automation}} ({{ICRA}})}, May 2020, pp. 7186--7192.

\bibitem{crosscombeRobustDistributedDecisionmaking2017}
M.~Crosscombe, J.~Lawry, S.~Hauert, and M.~Homer, ``Robust distributed decision-making in robot swarms: {{Exploiting}} a third truth state,'' in \emph{2017 {{IEEE}}/{{RSJ International Conference}} on {{Intelligent Robots}} and {{Systems}}}.\hskip 1em plus 0.5em minus 0.4em\relax Vancouver, BC, Canada: IEEE, 2017, pp. 4326--4332.

\bibitem{reinaDesignPatternDecentralised2015}
A.~Reina, G.~Valentini, C.~{Fern{\'a}ndez-Oto}, M.~Dorigo, and V.~Trianni, ``A design pattern for decentralised decision making,'' \emph{PLOS ONE}, vol.~10, no.~10, p. e0140950, Oct. 2015.

\bibitem{schmicklCollectivePerceptionRobot2007}
T.~Schmickl, C.~M{\"o}slinger, and K.~Crailsheim, ``Collective perception in a robot swarm,'' in \emph{Swarm {{Robotics}}}, E.~{\c S}ahin, W.~M. Spears, and A.~F.~T. Winfield, Eds.\hskip 1em plus 0.5em minus 0.4em\relax Berlin, Heidelberg: Springer Berlin Heidelberg, 2007, vol. 4433, pp. 144--157.

\bibitem{almansooriComparativeStudyDecision2022}
A.~Almansoori, M.~Alkilabi, and E.~Tuci, ``A comparative study on decision making mechanisms in a simulated swarm of robots,'' in \emph{2022 {{IEEE Congress}} on {{Evolutionary Computation}} ({{CEC}})}, Jul. 2022, pp. 1--8.

\bibitem{abdelliMaximumLikelihoodEstimate2023}
A.~Abdelli, A.~Yachir, A.~Amamra, and B.~Khaldi, ``Maximum likelihood estimate sharing for collective perception in static environments for swarm robotics,'' \emph{Robotica}, vol.~41, no.~9, pp. 2754--2773, Sep. 2023.

\bibitem{pfisterCollectiveDecisionmakingBayesian2022}
K.~Pfister and H.~Hamann, ``Collective decision-making with bayesian robots in dynamic environments,'' in \emph{2022 {{IEEE}}/{{RSJ International Conference}} on {{Intelligent Robots}} and {{Systems}} ({{IROS}})}, Oct. 2022, pp. 7245--7250.

\bibitem{pfisterCollectiveDecisionmakingChange2023}
------, ``Collective decision-making and change detection with bayesian robots in dynamic environments,'' in \emph{2023 {{IEEE}}/{{RSJ International Conference}} on {{Intelligent Robots}} and {{Systems}} ({{IROS}})}, Oct. 2023, pp. 8814--8819.

\bibitem{morlinoCollectivePerceptionSwarm2010}
G.~Morlino, V.~Trianni, and E.~Tuci, ``Collective perception in a swarm of autonomous robots,'' in \emph{Proceedings of the {{International Conference}} on {{Evolutionary Computation}}}.\hskip 1em plus 0.5em minus 0.4em\relax Valencia, Spain: {SciTePress - Science and and Technology Publications}, 2010, pp. 51--59.

\bibitem{strobelManagingByzantineRobots2018}
V.~Strobel, E.~C. Ferrer, and M.~Dorigo, ``Managing byzantine robots via blockchain technology in a swarm robotics collective decision making scenario,'' in \emph{Proceedings of the 17th {{International Conference}} on {{Autonomous Agents}} and {{MultiAgent Systems}}}, ser. {{AAMAS}} '18.\hskip 1em plus 0.5em minus 0.4em\relax Stockholm, Sweden: {International Foundation for Autonomous Agents and Multiagent Systems}, 2018, pp. 541--549.

\bibitem{strobelBlockchainTechnologySecures2020}
V.~Strobel, E.~Castell{\'o}~Ferrer, and M.~Dorigo, ``Blockchain technology secures robot swarms: A comparison of consensus protocols and their resilience to byzantine robots,'' \emph{Frontiers in Robotics and AI}, vol.~7, May 2020.

\bibitem{strobelRobotSwarmsNeutralize2023}
V.~Strobel, A.~Pacheco, and M.~Dorigo, ``Robot swarms neutralize harmful byzantine robots using a blockchain-based token economy,'' \emph{Science Robotics}, vol.~8, no.~79, p. eabm4636, Jun. 2023.

\bibitem{christensenExogenousFaultDetection2007}
A.~L. Christensen, R.~O'Grady, M.~Birattari, and M.~Dorigo, ``Exogenous fault detection in a collective robotic task,'' in \emph{Advances in {{Artificial Life}}}, ser. Lecture {{Notes}} in {{Computer Science}}, F.~{Almeida e Costa}, L.~M. Rocha, E.~Costa, I.~Harvey, and A.~Coutinho, Eds.\hskip 1em plus 0.5em minus 0.4em\relax Berlin, Heidelberg: Springer, 2007, pp. 555--564.

\bibitem{khadidosExogenousFaultDetection2015}
A.~Khadidos, R.~M. Crowder, and P.~H. Chappell, ``Exogenous fault detection and recovery for swarm robotics,'' \emph{IFAC-PapersOnLine}, vol.~48, no.~3, pp. 2405--2410, Jan. 2015.

\bibitem{taraporeFaultDetectionSwarm2019}
D.~Tarapore, J.~Timmis, and A.~L. Christensen, ``Fault detection in a swarm of physical robots based on behavioral outlier detection,'' \emph{IEEE Transactions on Robotics}, vol.~35, no.~6, pp. 1516--1522, Dec. 2019.

\bibitem{okeeffeFaultDiagnosisRobot2017}
J.~O'Keeffe, D.~Tarapore, A.~G. Millard, and J.~Timmis, ``Towards fault diagnosis in robot swarms: {{An}} online behaviour characterisation approach,'' in \emph{Towards {{Autonomous Robotic Systems}}}, Y.~Gao, S.~Fallah, Y.~Jin, and C.~Lekakou, Eds.\hskip 1em plus 0.5em minus 0.4em\relax Cham: Springer International Publishing, 2017, vol. 10454, pp. 393--407.

\bibitem{okeeffeFaultDiagnosisRobot2017a}
------, ``Fault diagnosis in robot swarms: {{An}} adaptive online behaviour characterisation approach,'' in \emph{2017 {{IEEE Symposium Series}} on {{Computational Intelligence}} ({{SSCI}})}, Nov. 2017, pp. 1--8.

\bibitem{aziziHierarchicalArchitectureCooperative2012}
S.~M. Azizi and K.~Khorasani, ``A hierarchical architecture for cooperative actuator fault estimation and accommodation of formation flying satellites in deep space,'' \emph{IEEE Transactions on Aerospace and Electronic Systems}, vol.~48, no.~2, pp. 1428--1450, Apr. 2012.

\bibitem{taheriCyberattackMachineinducedFault2024}
M.~Taheri, K.~Khorasani, I.~Shames, and N.~Meskin, ``Cyberattack and machine-induced fault detection and isolation methodologies for cyber-physical systems,'' \emph{IEEE Transactions on Control Systems Technology}, vol.~32, no.~2, pp. 502--517, Mar. 2024.

\bibitem{elwanyRealTimeEstimationMean2009}
A.~Elwany and N.~Gebraeel, ``Real-time estimation of mean remaining life using sensor-based degradation models,'' \emph{Journal of Manufacturing Science and Engineering}, vol. 131, no.~5, p. 051005, Oct. 2009.

\bibitem{liuStochasticFilteringApproach2020}
B.~Liu, P.~Do, B.~Iung, and M.~Xie, ``Stochastic filtering approach for condition-based maintenance considering sensor degradation,'' \emph{IEEE Transactions on Automation Science and Engineering}, vol.~17, no.~1, pp. 177--190, Jan. 2020.

\bibitem{liuNovelAlgorithmQuantized2023}
Y.~Liu, Z.~Wang, C.~Liu, M.~Coombes, and W.-H. Chen, ``A novel algorithm for quantized particle filtering with multiple degrading sensors: {{Degradation}} estimation and target tracking,'' \emph{IEEE Transactions on Industrial Informatics}, vol.~19, no.~4, pp. 5830--5838, Apr. 2023.

\bibitem{pinciroliARGoSModularParallel2012}
C.~Pinciroli, V.~Trianni, R.~O'Grady, G.~Pini, A.~Brutschy, M.~Brambilla, N.~Mathews, E.~Ferrante, G.~Di~Caro, F.~Ducatelle, M.~Birattari, L.~M. Gambardella, and M.~Dorigo, ``{{ARGoS}}: A modular, parallel, multi-engine simulator for multi-robot systems,'' \emph{Swarm Intelligence}, vol.~6, no.~4, pp. 271--295, Dec. 2012.

\bibitem{soaresKheperaIVMobile2016}
J.~M. Soares, I.~Navarro, and A.~Martinoli, ``The khepera {{IV}} mobile robot: {{Performance}} evaluation, sensory data and software toolbox,'' in \emph{Robot 2015: {{Second Iberian Robotics Conference}}}, L.~P. Reis, A.~P. Moreira, P.~U. Lima, L.~Montano, and V.~{Mu{\~n}oz-Martinez}, Eds.\hskip 1em plus 0.5em minus 0.4em\relax Cham: Springer International Publishing, 2016, pp. 767--781.

\bibitem{talamaliWhenLessMore2021}
M.~S. Talamali, A.~Saha, J.~A.~R. Marshall, and A.~Reina, ``When less is more: {{Robot}} swarms adapt better to changes with constrained communication,'' \emph{Science Robotics}, vol.~6, no.~56, p. eabf1416, Jul. 2021.

\bibitem{crosscombeImpactNetworkConnectivity2022}
M.~Crosscombe and J.~Lawry, ``The impact of network connectivity on collective learning,'' in \emph{Distributed {{Autonomous Robotic Systems}}}, F.~Matsuno, S.-i. Azuma, and M.~Yamamoto, Eds.\hskip 1em plus 0.5em minus 0.4em\relax Cham: Springer International Publishing, 2022, vol.~22, pp. 82--94.

\bibitem{casellaStatisticalInference2002}
G.~Casella and R.~L. Berger, \emph{Statistical Inference}, 2nd~ed.\hskip 1em plus 0.5em minus 0.4em\relax Australia ; Pacific Grove, CA: Thomson Learning, 2002.

\end{thebibliography}

\end{document}